\documentclass{article}
\usepackage{PRIMEarxiv}

\usepackage{graphicx}
\usepackage{array}
\usepackage{makecell} 
\usepackage{booktabs}
\usepackage{todonotes}
\usepackage{arabtex}
\usepackage{utf8}
\usepackage{color,soul}
\usepackage{soul} 
\usepackage[utf8]{inputenc}

\usepackage{mathptmx}
\usepackage{newtxtext,newtxmath}
\setcode{utf8}
\usepackage[utf8]{inputenc} 
\usepackage[T1]{fontenc}    
\usepackage{hyperref}       
\usepackage{url}            
\usepackage{booktabs}       
\usepackage{amsfonts}       
\usepackage{nicefrac}       
\usepackage{microtype}      
\usepackage{lscape}
\usepackage{longtable}
\usepackage{booktabs}
\usepackage{fancyhdr}       
\usepackage{makecell}
\usepackage{multirow}
\usepackage[acronym]{glossaries}
\usepackage[nonumberlist]{glossaries-extra}
\makeglossaries

\newacronym{AI}{AI}{Artificial Intelligence}

\newacronym{LLM}{LLM}{Large Language Model}

\newacronym{MCQ}{MCQ}{Multiple Choice Question}

\newacronym{API}{API}{Application Programming Interface}

\newacronym{BA}{BA}{Balanced Accuracy}

\newacronym{MAE}{MAE}{Mean Absolute Error}

\newacronym{ML}{ML}{Machine Learning}

\newacronym{NLP}{NLP}{Natural Language Processing}

\newacronym{BERT}{BERT}{Bidirectional Encoder Representations from Transformers}

\newacronym{DSM}{DSM}{Diagnostic and Statistical Manual of Mental Disorders}

\newacronym{CoT}{CoT}{chain-of-thought}

\newacronym{ZS}{ZS}{zero-shot}

\newacronym{FS}{FS}{few-shot}

\newacronym{SoTA}{SoTA}{state-of-the-art}

\newacronym{BDI}{BDI}{Beck Depression Inventory}

\newacronym{fMRI}{fMRI}{functional magnetic resonance imaging}

\newacronym{EEG}{EEG}{electroencephalogram}

\newacronym{ECG}{ECG}{electrocardiogram}

\newacronym{PHQ}{PHQ}{Patient Health Questionnaire}

\newacronym{PTSD}{PTSD}{Post-traumatic stress disorder}

\newacronym{EHR}{EHR}{Electronic Health Records}

\newacronym{RAG}{RAG}{Retrieval-Augmented Generation}
\usepackage{graphicx}       
\usepackage{ltcaption}
\usepackage{pdflscape}  
\usepackage{pgfplots}
\usepackage{filecontents}
\usepackage{pgfcalendar}
\usepackage{datetime}
\usepackage{svg}
\usepackage{xcolor}
\usepackage{mdframed}

\usepackage[utf8]{inputenc}

\pgfplotsset{compat=1.17}

\usetikzlibrary{patterns, shapes.arrows, decorations.pathreplacing}

\newcolumntype{A}{>{\small\raggedright\arraybackslash}p{5cm}}

\title{A Comprehensive Evaluation of Large Language Models on Mental Illnesses in Arabic Context }

\author{Noureldin Zahran$^\dagger$, Aya E. Fouda$^\dagger$, Radwa J. Hanafy$^{\dagger,\ddagger}$ and  Mohammed E. Fouda$^\dagger$ \\
$^\dagger$Compumacy for Artificial Intelligence solutions, Cairo, Egypt.\\
$^\ddagger$ Department of Behavioural Health- Saint Elizabeths Hospital, Washington DC, 20032.\\
fouda@compumacy.com
}

\begin{document}
\maketitle

\begin{center}
\textbf{ABSTRACT}
\end{center}

Mental health disorders pose a growing public health concern in the Arab world, emphasizing the need for accessible diagnostic and intervention tools. Large language models (LLMs) offer a promising approach, but their application in Arabic contexts faces challenges including limited labeled datasets, linguistic complexity, and translation biases. This study comprehensively evaluates 8 LLMs, including general multi-lingual models, as well as bi-lingual ones, on diverse mental health datasets (such as AraDepSu, Dreaddit, MedMCQA), investigating the impact of prompt design, language configuration (native Arabic vs. translated English, and vice versa), and few-shot prompting on diagnostic performance. We find that prompt engineering significantly influences LLM scores mainly due to reduced instruction following, with our structured prompt outperforming a less structured variant on multi-class datasets, with an average difference of 14.5\%. While language influence on performance was modest, model selection proved crucial: Phi-3.5 MoE excelled in balanced accuracy, particularly for binary classification, while Mistral NeMo showed superior performance in mean absolute error for severity prediction tasks. Few-shot prompting consistently improved performance, with particularly substantial gains observed for GPT-4o Mini on multi-class classification, boosting accuracy by an average factor of 1.58. These findings underscore the importance of prompt optimization, multilingual analysis, and few-shot learning for developing culturally sensitive and effective LLM-based mental health tools for Arabic-speaking populations.


\section{Introduction}

Mental health disorders, including depression, anxiety, PTSD, and suicide, are major global public health concerns, affecting millions of individuals worldwide. According to the World Health Organization (WHO), over 280 million people globally suffer from depression, and anxiety disorders impact approximately 301 million individuals \cite{who2023depression, who_anxiety}. Suicide remains a leading cause of death, with an estimated 703,000 deaths annually \cite{who_mental_health_report}. These conditions significantly burden healthcare systems, productivity, and quality of life, underscoring the need for accessible and effective mental health interventions.

The Arab world is facing an alarming mental health crisis, with high prevalence rates of stress, depression, and anxiety reported across the region. A 2020 study revealed that approximately 35\% of individuals in the MENA region frequently experience stress, while about 29\% report symptoms of depression. Stress levels are particularly high in Tunisia (53\%), Iraq (49\%), and Jordan (42\%), whereas lower but still concerning rates are observed in Egypt (27\%), Algeria (27\%), and Sudan (22\%). Similarly, depression is most prevalent among Iraqis (43\%), Tunisians (40\%), and Palestinians (37\%), with notable but lower rates in Algeria (20\%), Morocco (20\%), and Sudan (15\%)\cite{arab_barometer}.  

A 2022 survey conducted in Egypt involving 3,134 individuals (1,619 females and 1,515 males) further highlights the severity of the issue. The study reported a 43.5\% depression rate, with a significant gender disparity: 52.9\% of females and 33.4\% of males experienced depressive symptoms \cite{depression_egypt_study}.  

A 2017 World Health Organization (WHO) report found that approximately 5\% of the Eastern Mediterranean Region's population suffers from depression, while 4\% experience anxiety disorders. These figures align closely with global estimates, where 4.4\% of the population suffer from depressive disorders, and 3.6\% struggle with anxiety \cite{who2017depression_region}.

However, this escalating crisis is compounded by a critical shortage of trained mental health professionals. In the Arab region, Egypt serves as a stark example. According to a 2020 WHO report, Egypt has only 0.84 psychiatrists per 100,000 people and 0.86 psychologists per 100,000 people, with a total of 847 psychiatrists and 861 psychologists nationwide \cite{who2020mental_egypt}. These numbers fall significantly below global averages. A 2020 WHO study reported that globally, the median number of mental health workers is 13 per 100,000 population, with 1.7 psychiatrists and 1.4 psychologists per 100,000 people \cite{who2020mental_global}. This severe shortage of mental health professionals limits access to timely diagnosis and intervention, exacerbating the burden of untreated psychiatric disorders across the region.

The growing use of social media in the Arab world has opened new avenues for understanding and addressing mental health needs. Social media platforms allow individuals to express their struggles with depression, anxiety, and other mental health issues, often through posts containing linguistic markers indicative of psychological distress \cite{De_Choudhury_De_2014, GUNTUKU201743}. For instance, individuals with depression may exhibit increased use of first-person pronouns and negative emotional language, while those with anxiety often express worry, uncertainty, and references to physical symptoms such as fatigue or insomnia \cite{coppersmith-etal-2014-quantifying}.

While social media provides a rich source of real-time, large-scale data for detecting mental health conditions, it also presents challenges. Individuals may engage in self-diagnosis or rely on unverified information for guidance, which risks misinterpretation and further distress \cite{baklola2024mental}. These issues are compounded in regions where access to professional services is limited, making it critical to develop reliable tools for analyzing and supporting mental health through digital platforms.

Artificial intelligence (AI) has emerged as a transformative tool in mental health diagnostics and intervention. AI systems excel at processing large volumes of unstructured data to identify subtle patterns associated with psychiatric conditions. Early work in natural language processing (NLP) demonstrated the utility of linguistic analysis for detecting mental health markers, enabling tasks such as sentiment analysis and risk assessment \cite{le2021machine}.

The advent of LLMs, such as OpenAI's GPT-4 \cite{achiam2023gpt}, marked a significant leap forward in NLP. These models, trained on massive datasets, can understand and generate human-like text while capturing complex emotional cues, thematic patterns, and linguistic nuances. In mental health contexts, LLMs have shown promise in identifying signs of depression, anxiety, and suicidal ideation by analyzing user-generated text from social media \cite{calvo2017natural}. They can further support mental health care through empathetic conversational agents and scalable diagnostic tools, particularly in under-resourced settings \cite{fitzpatrick2017delivering}. An additional advantage of LLM-based diagnostics is interpretability. Unlike traditional machine learning (ML) models, which often function as black boxes, LLMs can provide justifications or reasoning for their outputs, making their predictions more transparent and trustworthy \cite{rudin2019stop}. This interpretability is crucial in clinical and mental health applications, where understanding the basis of a diagnosis can enhance user confidence and clinician adoption.

While studies on psychiatric evaluations using LLMs have been conducted on English social media posts \cite{llm4psy, xu2024mental}, to the best of our knowledge, no equivalent studies have evaluated general or Arabic-focused LLMs in this domain. The closest effort is BiMediX \cite{bimedix}, which aimed to bridge the psychiatric knowledge gap by developing a bilingual Arabic-English model. However, their evaluation was limited to specialized psychiatric questionnaires, such as MEDMCQA and PubMedQA, with no assessments conducted in a social media context. This lack of evaluations on Arabic social media posts highlights a significant gap, given the growing role of social media. 

Despite their potential, the development of LLMs for mental health diagnostics in Arabic faces several key challenges:  
\begin{itemize}  
    \item Lack of Annotated Arabic Datasets: The scarcity of labeled data for Arabic-language psychiatric tasks limits the development and evaluation of reliable models.  
    \item Linguistic Complexity: Arabic’s rich morphology, dialectal variation, and cultural context introduce significant challenges for model training and performance \cite{koubaa2024arabiangpt}.  
    \item Cross-Lingual Impact: Automatic translations from English to Arabic may introduce biases or inaccuracies, which can negatively affect diagnostic outcomes.  
\end{itemize}  
These challenges highlight the critical need for systematic evaluation of LLMs tailored to the Arabic context. Incorporating native Arabic datasets and translations is essential to better understand language-related biases and ensure accurate diagnostic performance.

This paper makes several key contributions to the field of AI-driven mental health diagnostics.
\begin{enumerate}
    \item It provides a comprehensive empirical assessment of multiple LLMs on Arabic mental health datasets, covering various psychiatric disorders and severity levels.
    \item It introduces a controlled experimental design to evaluate the impact of prompt engineering, including prompt structure, style, and few-shot prompting techniques.
    \item It examines the effects of language and translation on diagnostic accuracy by comparing native Arabic datasets to translations of English corpora.
    \item It identifies challenges specific to Arabic-language psychiatric tasks, offering novel insights for improving culturally sensitive AI tools in the Arab world.
\end{enumerate}

The remainder of this paper is organized as follows:  
\textbf{Section }\textbf{\ref{sec:datasets}} describes the data collection process, including the identification of both native Arabic datasets and translated English datasets used for evaluation.  
\textbf{Section }\textbf{\ref{sec:section 3}} details the experimental setup, outlining the task types, prompt templates, parsing techniques, sampling methods, and evaluation metrics employed in the study.  
\textbf{Section }\textbf{\ref{sec:section 4}} presents the results, including an analysis of invalid responses, prompt performance, model comparisons, and the effect of language on diagnostic accuracy. The performance of few-shot prompting is also evaluated in this section, providing a discussion of the key findings, highlighting dataset difficulty, model-specific performance, and the influence of language biases.  
Finally, \textbf{Section }\textbf{\ref{sec:section 5}} concludes the paper by summarizing the key contributions and suggesting future directions for improving the deployment of LLMs in psychiatric diagnostics for Arabic-speaking populations.

\section{Data Collection}
\label{sec:datasets}
The first step to conduct this assessment was to collect relevant datasets. 
Collecting standard peer-published datasets allowed us to score the \gls{LLM}'s performance on labeled data, 
which enables us to compare and analyze the performance of the models quantitatively. Through the remainder of this paper, an all-cap abbreviation will be used for each dataset name to ensure consistency. This section details the descriptions of each dataset used, with further statistics provided in Appendix Section \ref{sec:dataset-stats}.

\subsection{Native Arabic Datasets}
\label{sec:native_arabic_datasets}

The search for mental disorder related datasets began with a focus on native Arabic datasets. Due to the limited prior work in this field, no restrictions were placed on dialect or collection source, enabling a more diverse and comprehensive dataset collection and evaluation.

Given the sensitivity of mental diagnosis tasks, preference was given to academically published and manually annotated datasets. However, this was not always feasible, and datasets that did not fully meet these criteria were included to ensure a comprehensive and generalizable evaluation. Dataset quality can later be assessed through the collective judgment of LLMs, which may help identify potential anomalies. The findings from the collected datasets are summarized in Table \ref{tab:arabic_datasets}.

\subsubsection[Depression Corpus of Arabic Tweets (DCAT)]{Depression Corpus of Arabic Tweets (DCAT) \cite{DCAT}}

The DCAT dataset consists of 10,000 tweets, labeled as either positive or negative for depression, with 5,000 samples in each class. The authors describe the dataset as the first manually annotated dataset for Arabic depression; however, they did not provide information regarding the annotators’ expertise or the criteria used for annotation. The available description of the dataset is minimal, as it is limited to a submission on the Harvard Dataverse platform without an accompanying research paper or detailed documentation.

\subsubsection[Modern Standard Arabic Mood Changing and Depression Dataset (MCD)]%
{Modern Standard Arabic Mood Changing and Depression Dataset (MCD) \cite{MCD}}

The original MCD dataset contains 1,229 samples. After removing duplicates (127) and missing values (7), the final dataset comprises 1,094 samples.  This dataset is multi-class, containing 9 classes related to depression based on the PHQ-9 questionnaire. The class labels and their distribution are presented in Table \ref{tab:dataset-stats}. The dataset was initially created and labeled using an automated keyword-based approach for each class. However, the authors noted that the dataset was later reviewed for validity. Details regarding the experience or qualifications of those who manually validated the samples were not provided.

\subsubsection[ARADEPSU]{ARADEPSU \cite{ARADEPSU}}

The ARADEPSU dataset was collected from older Arabic sentiment analysis datasets spanning 2016 to 2022. Initially, the samples were annotated by annotators whose qualifications were not specified. The authors then reviewed each sample, resolving conflicts by prioritizing their own revisions.  

This dataset includes three classes: Depression, Depression with Suicidal Ideation, and Non-depression. For this study, only the test split, consisting of 5,053 samples was used. Tables \ref{tab:arabic_datasets} and \ref{tab:dataset-stats} provide an overview of the sample lengths and class distributions.

\subsubsection[CAIRODEP]{CAIRODEP \cite{CAIRODEP}}

The CAIRODEP dataset was compiled from seven different sources to ensure diversity. These sources included crowdsourcing (a custom volunteer form), forums and online platforms, Twitter, translations of the English Reddit mental health dataset using Google Translate, and neutral posts extracted from existing datasets.

The dataset initially contained 7,000 samples, with 3,600 labeled as depression and 3,400 as normal. However, our analysis identified and removed 209 duplicate entries to ensure data integrity for evaluation. Table \ref{tab:dataset-stats} summarizes the sample length statistics.

\subsubsection[Twitter-based Arabic Mental Illness (AMI)]{Twitter-based Arabic Mental Illness (AMI) \cite{AMI}}
This dataset was published with two splits: train and test. To ensure comparability with future studies (that may train on the train-split), only the test split was used, which contained 621 samples after cleaning.  

The dataset was released with very limited description. The only available information was provided in the abstract on the dataset download page, stating that an accompanying paper was yet to be published. It was mentioned that the dataset was manually annotated, but the experience of the annotators was not specified.  

This dataset is unique as it includes two features: one for the disorder class, which can be Insomnia, Depression, Stress, Anxiety, or Bipolar Disorder, and another indicating the presence or absence of the disorder. Introduces a limitation in handling posts with multiple co-occurring disorders. For example, a post may exhibit symptoms of both stress and depression; however, the dataset format restricts each post to a single disorder label.  

The absence of duplicate posts in the dataset indicates that no entry is associated with more than one disorder, reinforcing its structure as a collection of independent binary classification tasks. To process this dataset, it was split into five subsets, one for each disorder, with each subset treated as a separate binary classification task.

\subsection[Mental Disorder Egyptian Arabic Dialect (MDE)]{Mental Disorder Egyptian Arabic Dialect (MDE) \cite{MDE}}
The dataset was manually collected from various social media platforms, including Reddit, Sanvello, Twitter, Happify, and Facebook, targeting posts from individuals experiencing mental disorders. The collected posts included both English and Egyptian Arabic content, which were translated and standardized into Egyptian Arabic to ensure consistency.

The dataset contains a total of 1,800 records, evenly distributed across three mental health conditions: 600 records for depression, 600 for anxiety, and 600 for suicidal tendencies.

\subsection{Translated English Datasets}  
Despite extensive search efforts and relaxed dataset restrictions, it is clear from Table \ref{tab:arabic_datasets} that most accessible Arabic datasets are limited to depression disorder and are either binary or multi-class in nature, with no severity or multi-label datasets available.  

This limitation motivated us to translate six English datasets from our earlier English evaluation study \cite{llm4psy}. The translation was performed using Google Translate, and each of these datasets will be discussed in the upcoming subsections. Initially, Meta's NLLB \cite{NLLB} machine translation model was considered for this task. However, upon manual inspection, the outputs were found to contain hallucinations and inaccuracies, rendering them unsuitable for the study's requirements. Consequently, higher-resource alternatives, including Google Translate, GPT-4, and DeepL, were explored. Among these, Google Translate was selected due to its reliability and its ability to minimize the hallucinations commonly observed with generative models. Table \ref{tab:english_datasets} summarizes the translated English datasets included in this study. Class distributions and sample length statistics for the translated versions of these datasets are provided in Appendix Section \ref{sec:dataset-stats}.

\begin{landscape}

\begin{table}[]
\centering
\caption{Summary of Arabic Mental Health Datasets. A `?' indicates that the annotator was not mentioned, while `Manual (?)' denotes that the annotation process was conducted manually by annotators of unreported experience or expertise.}
\label{tab:arabic_datasets} 
\resizebox{1.35\textwidth}{!}{%
\begin{tabular}{@{}p{4.5cm}p{2.5cm}p{3cm}p{3.2cm}p{3.5cm}p{2.5cm}p{3.5cm}@{}} 
\hline
\textbf{Name} &
\textbf{Task} &
\textbf{Disorder(s)} &
\textbf{Source} &
\textbf{Annotator} &
\textbf{Sample Size} &
\textbf{Dataset Access} \\ 
\hline
DCAT  &
Binary &
Depression &
X (formerly Twitter)&
Manual (?) &
10,000 &
\href{https://doi.org/10.7910/DVN/YHMYEQ}{Publicly Available}\\

MCD &
Multi-Class &
\makecell[lt]{Suicidality\\+ 8 other classes}&
X &
Manual (?) &
1,229 &
\href{https://doi.org/10.17632/myrb2gky8w.1}{Publicly Available}\\

Alghamdi et al. (2020) \cite{9040556} &
Binary &
Depression &
Forum (Nafsany) &
Medical Experts &
1,642 &
\href{https://doi.org/10.1109/ACCESS.2020.2981834}{Inaccessible}\\

Almouzini et al. (2019) \cite{ALMOUZINI2019257} &
Binary &
Depression &
X &
Manual (?) &
2,722 &
\href{https://doi.org/10.1016/j.procs.2019.12.107}{Inaccessible}\\

Almars et al. (2022) \cite{cmc.2022.022609} &
Binary &
Depression &
X &
Medical Experts &
6,130 &
\href{https://doi.org/10.32604/cmc.2022.022609}{Inaccessible}\\

ARADEPSU &
Multi-Class &
\makecell[lt]{Depression and \\ Suicidal Ideation} &
X &
Manually by Authors &
20,213 &
\href{https://aclanthology.org/2022.wanlp-1.28.pdf}{Available Upon Request}\\

\makecell[lt]{Alabdulkreem et al. \\(2020) \cite{alabdulkreem2020}} &
Binary & 
Depression & 
X &
? &
10,000 &
\href{https://doi.org/10.1080/12460125.2020.1859745}{Inaccessible}\\

CAIRODEP &
Binary &
Depression &
\makecell[lt]{Crowdsourcing, Forums,\\ X, Translation} &
\makecell[lt]{Manual (Keywords)} &
7,000 &
\href{https://github.com/mramly/CairoMent}{Publicly Available}\\

ARANXIETY &
Severity &
Anxiety (3-point) &
X &
\makecell[lt]{Manual (AraPh Lexicon)} &
955 &
\href{https://doi.org/10.3389/fpsyg.2023.962854}{Inaccessible}\\

AMI &
Multi-Class &
\makecell[lt]{Insomnia, Depression,\\ Bi-polar, Anxiety} &
X &
Manual (?) &
3,500 &
\href{https://doi.org/10.17869/enu.2023.3027591}{Publicly Available}\\

MDE &
Multi-Class &
\makecell[lt]{Depression, Anxiety,\\ Suicidality} &
\makecell[lt]{Reddit, Sanvello,\\ X, Happify,\\Facebook} &
Manual (?) &
1,800 &
\href{https://github.com/esraa-mahmoudsaid/Classification-of-Mental-Disorders-in-Egyptian-dialect-of-Arabic?tab=readme-ov-file}{Inacessible} \\

MEDMCQA (BiMediX) &
MCQ &
- &
Translation &
Expert Reviewed &
4,180 &
\href{https://arxiv.org/pdf/2402.13253}{Publicly Available}
\\
\hline
\end{tabular}
}
\end{table}

\begin{table}[]
\centering
\caption{Summary of Selected English Mental Health Datasets} 
\label{tab:english_datasets} 
\resizebox{1.35\textwidth}{!}{%
\begin{tabular}{@{}p{3cm}p{3cm}p{3cm}p{3cm}p{3cm}p{2cm}p{3cm}@{}} \hline 
\textbf{Name} &
\textbf{Task} &
\textbf{Disorder(s)} &
\textbf{Source} &
\textbf{Annotator} &
\textbf{Sample Size} &
\textbf{Dataset Access} \\ \hline
DREADDIT &
\makecell[lt]{Binary} &
 Stress &
 Reddit &
 Crowdsourced &
 1,000 posts &
 \href{https://paperswithcode.com/paper/dreaddit-a-reddit-dataset-for-stress-analysis-1}{Publicly Available} \\
DEPSEVERITY &
\makecell[lt]{Severity} &
Depression (4-point)	 &
Reddit &
Medical Experts &
3,553 posts	 &
\href{https://github.com/usmaann/Depression_Severity_Dataset?utm_source=chatgpt.com0}{Publicly Available} \\
\makecell[lt]{SAD}  &
\makecell[lt]{Severity} &
\makecell[lt]{Daily Stressors \\ (9 categories)} &
\makecell[lt]{SMS-like conversations, \\ crowdsourced data, \\ LiveJournal} &
Crowdsourced  &
\makecell[lt]{6,850 \\ sentences} &
\href{https://github.com/PervasiveWellbeingTech/Stress-Annotated-Dataset-SAD}{Publicly Available} \\
DEPTWEET &
\makecell[lt]{Severity \\ Binary} &
Depression (4-point) &
X &
Crowdworkers (trained) &
40,191 tweets &
\href{https://github.com/mohsinulkabir14/DEPTWEET/tree/main/data}{Publicly Available}\\
SDCNL &
\makecell[lt]{Binary} &
\makecell[lt]{Depression, \\ Suicidal Ideation} &
Reddit (r/SuicideWatch, r/Depression) &
Automated &
1,895 posts &
\href{https://github.com/ayaanzhaque/SDCNL/blob/main/data/combined-set.csv}{Publicly Available} \\

RED SAM & 
  \makecell[lt]{Severity} &
  Depression (3-point) &
  Reddit &
  Crowdworkers  &
  16,632 posts  &
\href{https://raw.githubusercontent.com/ThejasHaridas/NLP_PROJECT_DEPRESSION/main/Augdata.csv}{Publicly Available} \\ \hline
\end{tabular}%
}

\end{table}

\end{landscape}

\subsubsection{DREADDIT\cite{turcan2019dreaddit}}  
DREADDIT is a stress analysis dataset created using Reddit posts collected through the PRAW \gls{API}. The dataset initially consisted of 187,444 posts from 10 subreddits, grouped into five domains: interpersonal conflict (abuse, social), mental illness (anxiety, \gls{PTSD}), and financial need (financial).  

The posts were segmented into five-sentence units for annotation. Crowdworkers on Amazon Mechanical Turk labeled each unit as “Stress,” “Not Stress,” or “Can’t Tell.” To ensure quality, segments with low annotator agreement or marked as “Can’t Tell” were excluded. This filtering resulted in a final dataset of 3,553 labeled segments.

\subsubsection{SDCNL \cite{haque2021deep}}  
SDCNL is a dataset developed to distinguish suicidal ideation from depression in social media posts, with the goal of improving the classification of these closely related mental health conditions. The dataset contains 1,895 posts sourced from two Reddit subreddits: r/SuicideWatch and r/Depression. This work addresses a significant research gap in differentiating between these two mental health states, which is crucial for tailoring appropriate interventions.  

The dataset was not manually labeled. Instead, an unsupervised method was employed to refine the initial labels, which were assigned based on the subreddit of origin. This approach enhanced the accuracy of the classifications without the need for direct human supervision.

\subsubsection{SAD \cite{mauriello2021sad}}  
The SAD dataset is designed to identify daily stressors in SMS-like conversations, with the goal of improving chatbot interactions by classifying stressors in text. It contains 6,850 sentences labeled into nine categories, sourced from stress management literature, chatbot conversations, crowdsourcing, and LiveJournal.

The categories are as follows: Work (1,043), School (1,043), Financial Problems (152), Emotional Turmoil (238), Social Relationships (99), Family Issues (101), Health Issues (103), Everyday Decisions (101), and Other (123). Sentences were labeled and rated for stress severity on a 10-point scale by crowd workers. The dataset, which includes stress indicators and metadata, is publicly available on GitHub for research purposes.

\subsubsection{DEPTWEET \cite{kabir2023deptweet}}  
DEPTWEET is a dataset designed to detect depression severity levels in social media posts. It consists of 40,191 tweets categorized into four severity levels: non-depressed, mild, moderate, and severe. Tweets were collected using depression-related keywords and annotated by 111 trained crowdworkers following guidelines based on \gls{DSM}-5 and \gls{PHQ}-9 assessment tools.  

Each tweet was labeled by at least three annotators, with the final labels determined through majority voting. A confidence score was assigned to indicate annotator agreement. The dataset also includes metadata such as tweet ID, reply count, retweet count, and like count, facilitating further analysis.

\subsubsection{RED SAM \cite{sampath2022data, kayalvizhi2022findings}}  
RED SAM is a dataset developed to detect depression severity levels in social media posts. It contains 16,632 Reddit posts categorized into three classes: "Not Depressed", "Moderately Depressed", and "Severely Depressed." The dataset aims to assess both the presence and severity of depression to support better treatment and intervention strategies.  

Posts were collected from mental health-focused subreddits, including "r/MentalHealth", "r/depression”, "r/loneliness", "r/stress", and "r/anxiety”. Annotation was conducted by two domain experts using guidelines that evaluated factors such as post length, emotional expression, and specific language patterns indicative of depression severity.

\section{Experimental Setup}  
\label{sec:section 3}
The experimental process began with identifying the disorders and task types present in the collected datasets. Based on this analysis, suitable prompt templates, parsers, and evaluation metrics were designed. Sampling methods were applied where applied according to each dataset's type, and the selected models were tested accordingly. The following sections outline each step in detail.

\subsection{Task Types}

\textbf{Binary classification}: Datasets with this task type have a single label for each sample/entry. 
This label is either 0 or 1, such as "depressed" or 
"normal." The purpose of binary tasks is to measure 
how well the model can distinguish between two opposing 
categories. For example, the model must decide whether 
a user is experiencing symptoms of depression or not. 
In mental health contexts, binary classification tasks 
are essential because they enable the early detection of 
conditions like depression by flagging users who fall 
into a specific category of interest, such as "depressed."

\textbf{Severity classification}: Datasets with this 
task type have a single label that is a discrete numeric 
value representing the severity of a specific disorder. 
The range of values is dataset-dependent, and these 
numerical values are typically associated with a textual 
representation. For example, "Mild Depression" might be 
mapped to 1. This task can be thought of as a 
classification problem within a single disorder, 
where the goal is to classify how severe an individual’s 
disorder is. The purpose of severity tasks is to measure 
how well the model can assess the intensity or progression 
of a particular mental health issue, like distinguishing 
between mild and severe cases of depression, allowing for 
more nuanced treatment and intervention.

\textbf{Multiclass classification}: In multi-class tasks, one label can take 
on multiple potential classes. A challenging example would be datasets 
such as MDE, where classes include "Suicidality," "Depression," and 
"Anxiety," or SDCNL, where labels are "Depression" and "Suicide." This 
task type is tricky because the model must distinguish between disorders 
that often have overlapping symptoms. The model must learn to identify 
subtle differences between conditions like anxiety and depression, which 
is critical for delivering accurate mental health diagnostics.

\textbf{MCQ (Multiple Choice Questions)}: The only instance of this 
task type in our dataset is MedMCQA. While it has a single label, this 
label takes on different values depending on the sample. This is not 
equivalent to a traditional multi-class problem, as each sample represents 
a question with a different set of possible answers. The purpose of this 
task is to test the model’s ability to understand and answer medical 
questions in a multiple-choice format, which requires a deeper 
understanding of the context and underlying medical knowledge.

\subsection{Prompt Templates}

For each of the aforementioned tasks, exactly two prompt templates were designed. These templates are generalized, meaning they are not tailored to a specific disorder but follow a rigorous structure with formattable variables for the disorder name that seamlessly integrate into the sentence. For example, in a prompt like "Is this person showing symptoms of X?", X could be \{Depression, Suicidal ideation, Stress, etc.\}. This approach ensures that the instruction remains consistent across all disorders and datasets, guaranteeing fair comparisons while also simplifying the experimental setup.

Additionally, the templates are generalized across tasks, meaning all tasks share the same base structure with slight modifications to adjust the output formatting depending on the task requirements. For instance, binary classification prompts expect a "yes" or "no" response, multi-class prompts require selecting from a provided list, and severity prompts request a severity rating within a specified range. To introduce few-shot prompting, a section containing sample-input and ground-truth pairs is simply appended to the template before the original post that the model is expected to diagnose. This straightforward modification ensures flexibility while retaining consistency across all tasks.

Based on earlier English evaluation work \cite{llm4psy}, it was observed that even subtle modifications to a prompt—such as changing or repeating words—could unexpectedly lead to significant performance variations. Consequently, it was concluded that over-engineering prompts is counterproductive, as minor adjustments may yield better results without clear justification. This phenomenon, often referred to as a "random search problem," is particularly relevant because different models can respond unpredictably to such minor prompt changes. Therefore, for each task, two reasonable and well-structured prompts were designed to cover the necessary instructions, including context, the question, constraints, and guidelines.

For each task, two prompt templates were systematically designed to ensure fair comparisons and to analyze the models' sensitivity to prompt variations. While both prompts maintain semantically identical content and instructions, they differ in structure and contextual framing. Prompt style 1 is less structured and explicitly instructs the model to "act as a psychologist" when generating its response. In contrast, prompt style 2 follows a more structured format but omits the "act as a psychologist" clue. Later on in the paper prompt 1 and prompt 2 (in their zero-shot form) are referred to as ZS-1 and ZS-2. The few-shot experiment section \ref{sec:fewshot} details the changes made to convert from zero-shot to few-shot setting.

This design choice serves two purposes. First, it allows for an evaluation of the models' responses to changes in prompt structure, as even subtle modifications can lead to significant performance variations. Second, it enables an analysis of the models' behavior when the contextual framing provided by the psychologist role is removed. By comparing the two styles, this setup reveals potential style preferences and shows how the absence of role-based context impacts the model's output.  

Since the binary classification prompt templates used in the earlier English evaluation satisfied the criteria, two of these templates were selected and translated into Arabic using GPT-4o, with subsequent manual corrections. The final set of prompts is provided in the Appendix (Section \ref{sec:all_prompt_templates}). Figures \ref{fig:fs-p} and \ref{fig:fs-p_1} present an example of ZS-2 formatted for binary depression classification in Arabic and English, respectively.


\begin{figure}[htb]
\begin{mdframed}[backgroundcolor=gray!10, linecolor=gray!10]
    \centering
    \textbf{Example Arabic Prompt (ZS-2 for Binary Depression Classification)}
\begin{tikzpicture}
    \node[fill=white!5, draw=gray, rounded corners, inner sep=10pt, text width=0.90\textwidth] (box1) {
    \vspace{2mm}
    {\small 
    \begin{RLtext} \textbf{[المهمة]} \end{RLtext} \\
    \begin{RLtext} حلل منشور التواصل الاجتماعي التالي لتحديد ما إذا كان الكاتب يُظهر أعراضًا واضحة لـ الاكتئاب وفقًا للإرشادات المقدمة.\end{RLtext} 
    \begin{RLtext} \textbf{[للإرشادات]} \end{RLtext} 

    \begin{RLtext}- إجابة موجزة: أجب بـ نعم فقط (يُظهر أعراضًا واضحة لـ الاكتئاب) أو لا (لا يُظهر أعراضًا واضحة لـ الاكتئاب).\end{RLtext}

    \begin{RLtext} - عدم تقديم تفسيرات؛ لا تقدم تفسيرات لتقييمك .\end{RLtext} 
    \begin{RLtext} -  الغموض: إذا كان المنشور غير واضح، فاختر التسمية الأكثر احتمالاً.\end{RLtext} 

      \begin{RLtext} \textbf{[المنشور]} \end{RLtext} 

    \begin{RLtext} \textbf{{هنا يأتي نص المنشور}} \end{RLtext} 
    }
    };
\end{tikzpicture}
    \caption{Example of a formatted prompt (ZS-2) for binary depression classification.}
    \label{fig:fs-p}
\end{mdframed}
\end{figure}

\begin{figure}[htb]
\begin{mdframed}[backgroundcolor=gray!10, linecolor=gray!10]
    \centering
    \textbf{Example English Prompt (ZS-2 for Binary Depression Detection)}
    \begin{tikzpicture}
        \node[fill=white!5, draw=gray, rounded corners, inner sep=10pt, text width=0.90\textwidth] (box1) {
        \textbf{[Task]}\\
        Analyze the following social media post to determine if the writer exhibits clear symptoms of depression according to the provided guidelines.\\
       \vspace{2mm}
        \textbf{[Guidelines]}\\
        \vspace{2mm}

        - Concise Response: Respond only with 'Yes' (exhibits clear symptoms of depression) or 'No' (does not exhibit clear symptoms of depression).\\
               \vspace{2mm}
        - No Explanations: Do not provide explanations for your assessment.\\
        - Ambiguity: If the post is unclear, choose the most probable label.\\
       \vspace{2mm}
        \textbf{[Post]}\\
        };
    \end{tikzpicture}
    \caption{Example of a formatted prompt (ZS-2) for a binary classification task on depression.}
    \label{fig:fs-p_1}
\end{mdframed}
\end{figure}

\subsection{Parsing}
One critical step that plays a significant role in the computed metric score is how the models’ responses were parsed. Parsing the output incorrectly can lead to dramatic changes in evaluation outcomes. To simplify the parsing process, a constraining clue was introduced in the instruction prompt, explicitly asking the model to return a single word. This approach aims to ensure that parsing remains straightforward. Ideally, consistent adherence to this constraint would guarantee successful parsing by limiting the response to only the diagnosis. However, as further analyzed in the discussion, this is not always the case. Certain models occasionally disregard the instruction, producing more complex responses that complicate parsing.

In such cases, there is no guarantee that the parsed diagnosis accurately reflects the model's intended diagnosis. Therefore, our parsing methodology attempts to minimize such errors. The paragraphs below describe the parsing process for different task types:

\textbf{Binary Classification Parser}: The logic for the binary parser is straightforward. If the word "yes" (in Arabic) is found in the model’s response, the label is assigned as "yes." If the word "no" (in Arabic) is found, the label is assigned as "no." However, if both words are found or neither word is detected, the response is labeled as invalid.

\textbf{Severity Parser}: For the severity task, regular expressions (regex) were used to search for a pattern that contains a numerical value. If a valid number is found within the specified range for the dataset, it is assigned as the severity label. If the number is out of range, a floating point number, or if no match is found, the response is labeled as invalid.

\textbf{Multiclass Parser}: In multi-class classification tasks, the response was checked for any of the disorders listed in the instruction prompt. If no match is found, the response is labeled as invalid.

\textbf{Knowledge Parser}: For the knowledge-based multiple-choice questions, the search focused on identifying Arabic alphabetical choice letters equivalent to "a, b, c, d." These letters were required to remain standalone and not form part of a word. Formats such as (a), (A), or A) were considered valid for this purpose.

By applying these parsing methods, the goal was to minimize errors and ensure that the model's output adhered closely to the intended format. Nevertheless, as discussed later, parsing remains a potential source of error, particularly when models fail to strictly follow the prompt structure.

\subsection{Sampling}

Since all testing was conducted via \gls{API}s, evaluation costs had the potential to escalate significantly, particularly when running multiple models across large datasets. To mitigate these costs, two key cost-saving strategies were employed. First, only the test splits of datasets that were already pre-split were utilized. Second, for datasets exceeding 1,000 examples, \textbf{fair random sampling} was implemented to ensure balanced representation of all classes within the dataset. This approach allowed for the evaluation of up to 1,000 instances per dataset while preserving fairness across class distributions. The sample size was chosen based on Hoeffding's inequality (Equation \ref{eqn:hoeffding}), which provides an upper bound for the error difference between a sample and the full dataset, ensuring that the sample is sufficiently representative.

For datasets with severity metrics, such as the DEPSEVERITY and SAD datasets, the labels were binarized by assigning a "False" label (0) to posts or users with the minimum severity score and a "True" label (1) to those with any score above the minimum. This approach simplified the classification task while retaining meaningful distinctions between varying severity levels.

\begin{equation}
\label{eqn:hoeffding}
P\left( \left| \nu - \mu \right| > \varepsilon \right) \leq 2e^{-2\varepsilon^2 N}
\end{equation}
where \(\nu\) is the empirical mean (sample average), \(\mu\) is the true mean (population average), \(\varepsilon\) represents the deviation from the mean, and  \(N\) is the number of samples.

By substituting \(N = 1000\) and \(\varepsilon = 0.05\) into Equation \ref{eqn:hoeffding}, which represents a maximum discrepancy of 5\%, the probability of the in-sample and out-of-sample error deviating by more than 5\% is calculated to be 0.0135. This result implies a greater than 98.5\% probability that the error will remain within a 5\% margin of the true error, ensuring confidence in the fairness and representativeness of the sample.

\subsection{Evaluation Metrics}
\label{subsec:evaluation_metrics}

This study evaluated the performance of \gls{LLM}s using accuracy-based metrics, with Balanced Accuracy (\gls{BA}) serving as the primary measure across tasks. \gls{BA} was chosen because it provides a fair evaluation of model performance, particularly in the presence of class imbalances, by calculating the average recall for each class. This makes it a more reliable metric compared to traditional accuracy when dealing with uneven data distributions.

In many datasets, fair random sampling was applied to ensure balanced class distributions, allowing for a straightforward and equitable comparison of performance across all classes. However, in cases where fair random sampling was not feasible—such as with datasets like SAD, which exhibited sparse class distributions—random sampling was employed instead. Despite this, \gls{BA} was still used as the evaluation metric because it inherently accounts for class representation and adjusts for imbalances naturally, even when the data is not explicitly balanced during sampling.

By using \gls{BA}, the study ensures consistent and equitable evaluation of models across diverse datasets and task types.

\paragraph{Balanced Accuracy (BA)}
Balanced Accuracy is calculated as the average recall across all classes:
\[
\text{BA} = \frac{1}{C} \sum_{c=1}^{C} \frac{TP_c}{TP_c + FN_c},
\]
where \(C\) is the number of classes, \(TP_c\) is the number of true positives for class \(c\), and \(FN_c\) is the number of false negatives for class \(c\). This formula ensures that the metric is not biased toward classes with more samples, providing a fair assessment of the model's performance across the entire dataset.

\paragraph{Mean Absolute Error (MAE)}
For the disorder severity task, two metrics were used: \gls{BA} and Mean Absolute Error (\gls{MAE}). While \gls{BA} measures how often the model correctly predicts the exact severity level, \gls{MAE} focuses on the average size of the prediction errors. To ensure consistency across datasets with different severity scales, a normalized version of \gls{MAE} was used:
\[
\text{MAE}_{\text{norm}} = \frac{1}{N} \sum_{i=1}^{N} \frac{|\hat{y}_i - y_i|}{L-1},
\]
where \(N\) is the number of samples, \(\hat{y}_i\) is the predicted severity level for sample \(i\), \(y_i\) is the actual severity level, and \(L\) is the total number of severity levels in the dataset. Normalizing by \(L\) ensures the metric remains comparable across datasets with different scales. When referring to \gls{MAE} in the remainder of this paper, it specifically denotes the normalized version calculated using the equation stated above.

Together, \gls{BA} and \gls{MAE} offer a complementary view of model performance. While \gls{BA} highlights how often the model achieves exact predictions, \gls{MAE} provides insights into how far off the predictions are when they are not exact. This combination of metrics allows for a more complete evaluation of \gls{LLM} capabilities.

\subsection{Models}

We selected eight models for this evaluation: \textit{Gemma 2 9B}, \textit{GPT-4o Mini}, \textit{Llama 3 70B}, \textit{Mistral NeMo}, \textit{Phi-3.5 Mini}, \textit{Phi-3.5 MoE}, \textit{Aya 32B}, and \textit{Jais 13B}. The models were chosen based on their strong general performance observed in our previous English evaluation study and their computational feasibility for further fine-tuning. Aya 32B and Jais 13B were included after a preliminary evaluation of Arabic-focused models, which is detailed in Appendix~\ref{sec:arabic_models}.

A summary of these models, including their parameters, sources, and API providers, is presented in Table \ref{tab:models}.  

The primary reason for selecting relatively smaller models is the feasibility of fine-tuning them. While larger models often yield better out-of-the-box performance, this study aims to thoroughly evaluate these models, identify the best-performing one. A later study will then proceed with fine-tuning a selected model based on performance to further improve it in Arabic mental health-related tasks. 

\renewcommand{\arraystretch}{1.2} 
\begin{table}[htbp]
\centering
\caption{Comparison of the different models included in the study, their parameters, source, and API provider}
\label{tab:models}
\begin{tabular}{l|l|l|l}
\hline
Model        & \# Parameters & Source     & API Used \\ \hline
GPT-4o Mini  & Proprietary   & OpenAI     & OpenAI            \\
Llama 3 70B  & 70 Billion    & Meta       & Groq              \\
Gemma 2 9B   & 9 Billion     & Google     & Groq              \\
Mistral NeMo & 12 Billion    & Mistral AI & Mistral AI        \\
Phi-3.5 Mini & 3.8 Billion   & Microsoft  & Azure       \\
Phi-3.5 MoE  & \begin{tabular}[c]{@{}l@{}}6.6 Billion Active - 42 Total\end{tabular} & Microsoft & Azure \\
Jais & 13 Billion    & InceptionAI & HuggingFace  \\
Aya & 32 Billion    & Cohere & Cohere  \\
\hline
\end{tabular}
\end{table}

\section{Results \& Discussion}
\label{sec:section 4}
This section presents the results and findings of the evaluation. The presentation begins with an analysis of invalid responses, followed by a detailed examination of prompt performance and a general analysis (Sections \ref{sec:invalid_response_analysis}, \ref{sec:prompt_performance_analysis}, and \ref{sec:general_analysis}). These sections provide comprehensive results, serving both as a summary and as a foundation for subsequent, more focused analyses.

Later sections offer additional insights into the impact of language and translation on model performance. Specifically, Section \ref{sec:english_to_arabic_analysis} evaluates the effect of translating English datasets into Arabic, comparing the results when models are evaluated using prompts in languages that match the dataset language \ref{sec:native-arabic-vs-enlgish-translated} explores the reverse process, where native Arabic datasets are translated into English and evaluated using Arabic prompts. However, the prompts here are fixed to be in Arabic. These experiments aim to isolate the effect of language on diagnostic accuracy while providing further analysis beyond the initial results.

For clarity, the translation experiments are incorporated into the earlier discussed heatmaps and figures by expanding the dataset axis. Translated English datasets are labeled as \textit{"(AR) {English Dataset Name}"}, indicating their evaluation in the Arabic configuration. Conversely, translated Arabic datasets are labeled as \textit{"(EN) {Arabic Dataset Name}"}, denoting their evaluation in the English configuration. This systematic inclusion ensures that the findings are presented cohesively, allowing for cross-language and cross-model comparisons under consistent evaluation conditions.

When comparing the performance of two experiments or configurations, this study adopts a clear convention for reporting results. Differences are represented as absolute values, without the use of a percentage sign, even when referring to metrics such as Balanced Accuracy. For example, if Model X outperforms Model Y by a value of 10, this implies \textit{Performance X - Performance Y = 10}, with no percentage symbol included. In contrast, relative improvements are reported as ratios and expressed as percentages. Under this convention, a reported improvement of 10\% signifies that \textit{Performance X = 1.1x Performance Y}. This distinction ensures clarity when interpreting the results, particularly when comparing absolute differences and relative gains across models, prompts, and experimental configurations.

\subsection{Invalid Response Analysis}
\label{sec:invalid_response_analysis}

Before presenting the results, it is important to examine how well the responses could be parsed. Metrics are only meaningful if the parsing process successfully categorized the outputs into valid diagnostic classes. Additionally, the overall performance of the models is inherently limited by the percentage of responses that were parsed correctly.

\begin{figure}[htbp]
    \centering
    \includegraphics[width=1\textwidth]{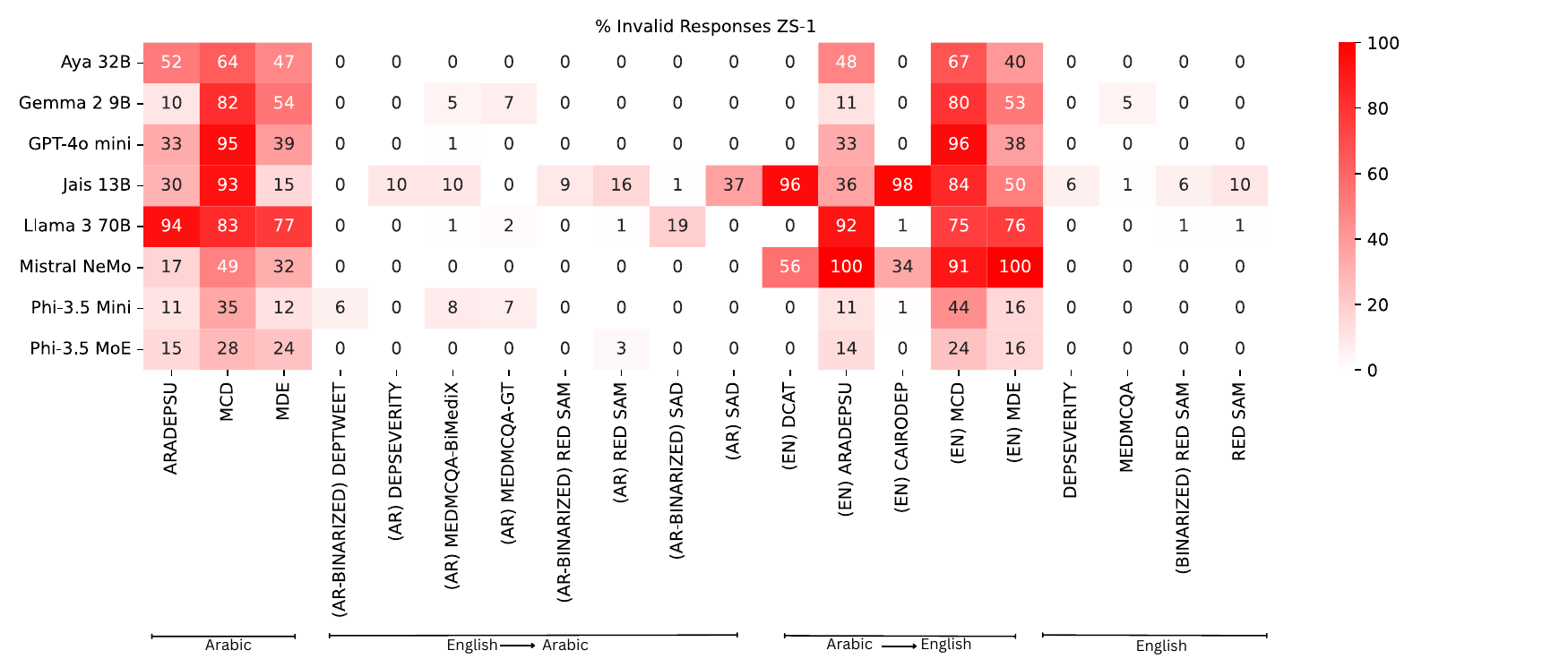} %
        \caption{Percentage of invalid responses for each dataset-model trial using the ZS-1 Prompt. Columns with a maximum invalid response percentage below 5\% were omitted for clarity.}

    \label{fig:zs1_invalid_response}          
\end{figure}

\begin{figure}[htbp]
    \centering
    \includegraphics[width=1\textwidth]{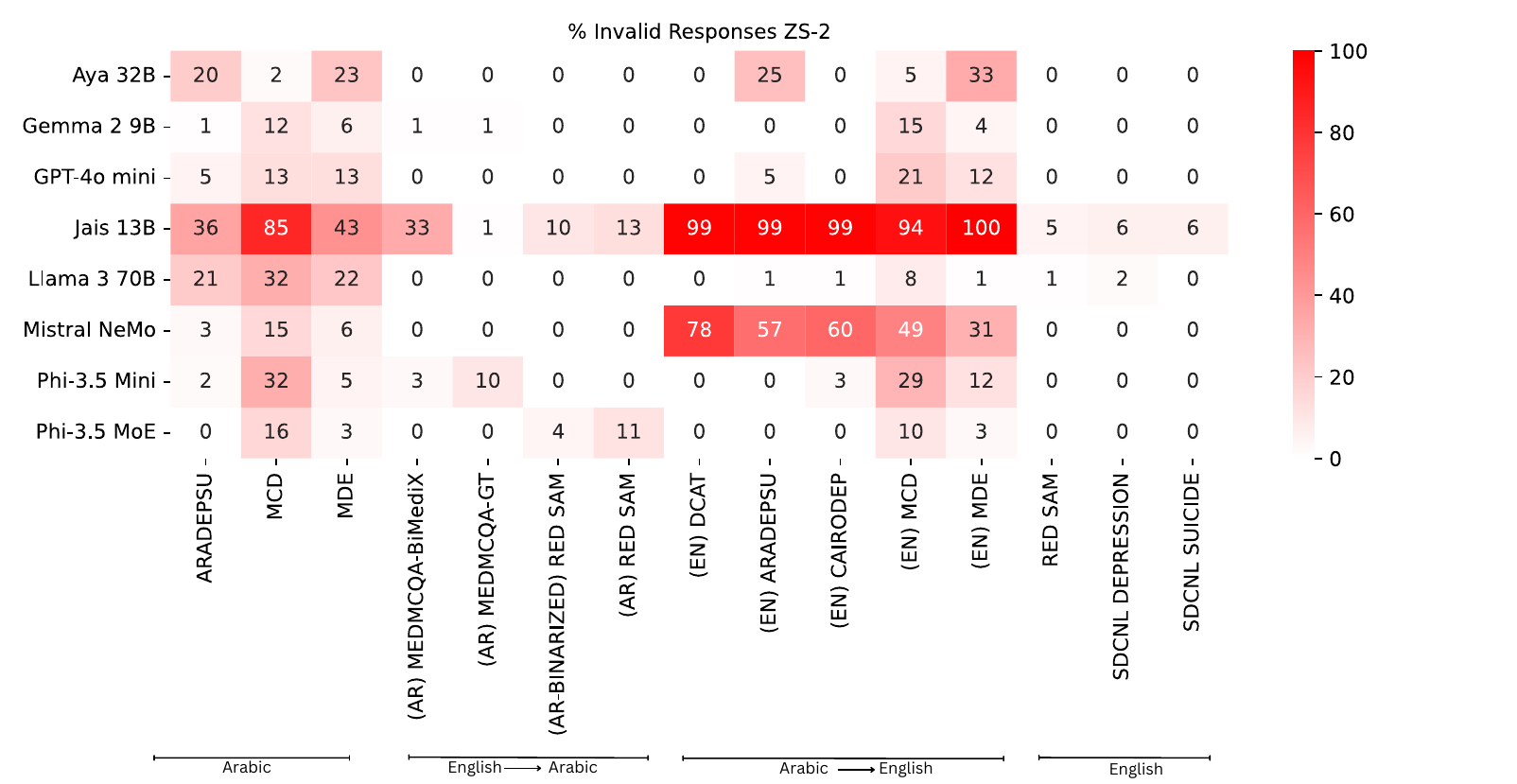}  %
        \caption{Percentage of invalid responses for each dataset-model trial using the ZS-2 Prompt. Columns with a maximum invalid response percentage below 5\% were omitted for clarity.}
    \label{fig:zs2_invalid_response}          
\end{figure}

This subsection focuses on parsing validity, which measures how many outputs could be classified into a diagnosis category. Figures \ref{fig:zs1_invalid_response} and \ref{fig:zs2_invalid_response} show the percentage of invalid responses for Prompts ZS-1 and ZS-2.

From these figures, it can be observed that most models generally returned valid responses across all severity, binary, and knowledge datasets. However, for multi-class datasets, only ZS-2 achieved consistent performance overall, with the exception of Jais 13B, which failed to produce instruction-following responses on the English-translated version of the Arabic datasets. Similarly, Mistral NeMo encountered issues in these experiments. In contrast, ZS-1 demonstrated significantly poorer performance across models on multi-class datasets, including both native and translated versions.

While a higher rate of invalid responses was expected for multi-class prompts compared to binary or severity datasets—due to the added complexity of providing the model with a list of disorders and requiring it to identify the most relevant one. The markedly poor performance of ZS-1, along with the observed disparity between ZS-1 and ZS-2, was beyond initial expectations. This issue is explored further in Appendix \ref{sec:response_length_analysis}.

None of the experiments conducted on FS-2 exceeded 10\% invalid responses, except for the combination of Phi-3.5 MoE and MDE, which recorded 16\%. A notable reduction in invalid responses was observed across the remaining experiments, particularly for GPT-4o Mini.

\subsection{Prompt Performance Analysis}
\label{sec:prompt_performance_analysis}

\begin{landscape}
\begin{figure}[htbp]
\centering
\includegraphics[width=1.6\textwidth]{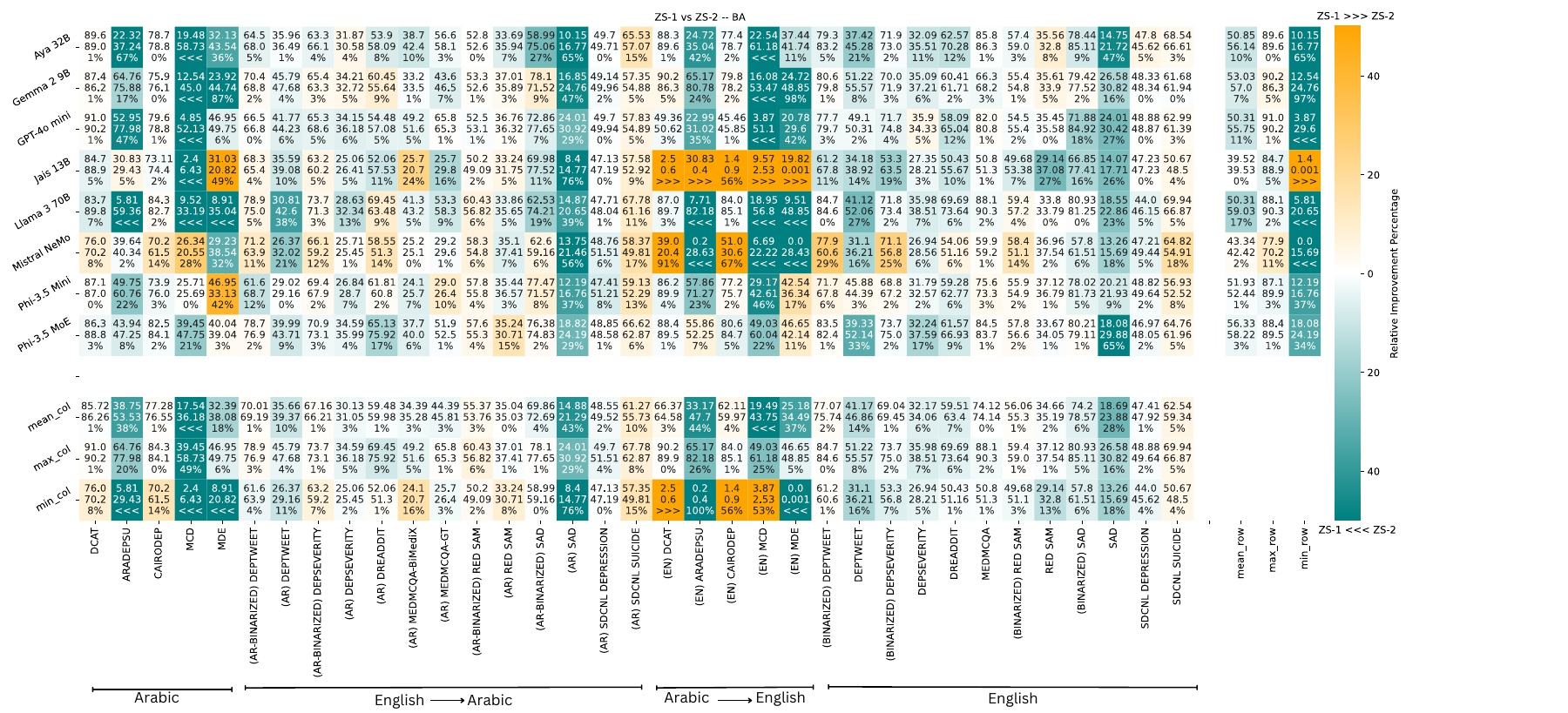}  %
\caption{Comparison of performance of models on BA for prompts ZS-1 and ZS-2 across all datasets, measuring relative improvement percentage.}
\label{fig:zs1vszs2_ba_all}
\end{figure}
\end{landscape}

This section presents the results of the experimentation on the Balanced Accuracy metric, summarized in a heatmap (Fig. \ref{fig:zs1vszs2_ba_all}). The heatmap visually compares ZS-1 and ZS-2 across various tasks. Each cell in the heatmap displays three values: the accuracy of ZS-1 (top), the accuracy of ZS-2 (middle), and the relative improvement percentage when transitioning from ZS-1 to ZS-2 (bottom).

\[
\text{Relative Improvement Percentage}_\text{BA} = \left( \frac{\text{Higher BA}}{\text{Lower BA}} - 1 \right) \times 100
\]

Colors are used to emphasize improvements:orange highlights cases where ZS-1 outperformed ZS-2, while teal indicates the opposite. For example, on the Gemma 2 9B - ARADEPSU dataset, ZS-2 achieved an accuracy of 75.9\%, compared to 64.8\% for ZS-1. This represents a relative improvement of 17\%, or 1.17x better performance with ZS-2. This relative gain highlights differences caused by the prompts, independent of model performance and dataset difficulty.

Upon analyzing the figure, it is evident that ZS-2 generally outperformed ZS-1 across most model-dataset pairs. ZS-1 struggled significantly with multi-class datasets, which is expected because the models tended to produce a higher number of invalid responses on the multi-class datasets when using prompt ZS-1, as noted in earlier analyses.

To quantify performance differences, the number of instances where one prompt outperformed the other was counted. Only cases with a relative improvement greater than 5\% were considered to minimize the impact of random fluctuations. Based on this criterion, there were 45 cases where ZS-1 \texttt{>} ZS-2, 108 cases where ZS-1 \texttt{<} ZS-2, and 127 cases categorized as neutral or random fluctuations. Subtracting the performance of ZS-2 from ZS-1 and taking the average, it's found that performance increased by an average of 3 as a result of using ZS-2. The performance on binary datasets wasn't affected (<1 difference), while on severity datasets ZS-2 showed slightly higher BA (higher by 3), more noticeably the performance on multi-class increased by 14.5. However calculating these statistics on only the intersection of responses where both prompts led to a parsable response, it was found that the difference was only 1.6 in favor of ZS-1, asserting that the the improvement was only in-terms of instruction following.

To study the effect of the contextual clue, the inclusion of the “as a psychologist…” sentence in the prompt, an error analysis of binary dataset predictions was conducted. Confusion matrix (Table \ref{tab:zs1zs2_conf_delta}), indicates that this prompt variation increased the proportion of negative predictions (TN + FN) by 2.8\%. Similarly, considering only the Arabic datasets (native and translated), the proportion of negative predictions increased by 4.7\%. This suggests that the models became more cautious and were less likely to produce positive diagnoses.

\begin{table}[h]
\centering
\caption{Normalized aggregate confusion matrix (over all models and binary datasets) for ZS-1 subtracted by the aggregate confusion matrix for ZS-2, showing the differences in prediction errors.}
\label{tab:zs1zs2_conf_delta}
\begin{tabular}{lrrr}
\toprule
 & Predicted Invalid & Predicted False & Predicted True \\
\midrule
- & 0.000 & 0.000 & 0.000 \\
Actual False & -0.002 & 0.018 & -0.016 \\
Actual True & 0.000 & 0.010 & -0.010 \\
\bottomrule
\end{tabular}
\end{table}

To further investigate this observation, the analysis was extended to the error types in the severity datasets. Errors were categorized into two types: E1, representing overestimation of the disorder’s severity (Predicted severity > Ground Truth), and E2, representing underestimation (Predicted severity < Ground Truth). The results revealed that, for ZS-1, E1 errors accounted for 93.5\% of the total prediction errors, whereas for ZS-2, this percentage decreased to 88.8\%. These findings differ from the earlier analysis of binary datasets, suggesting opposite behavior on the severity datasets, where ZS-1 caused predictions to become more pessimistic. A similar trend was observed for Arabic datasets, with E1 errors increasing from 86.6\% on ZS-2 to 92.6\% on ZS-1.

\begin{figure}[h]
    \centering    
    \includegraphics[width=0.8\textwidth]{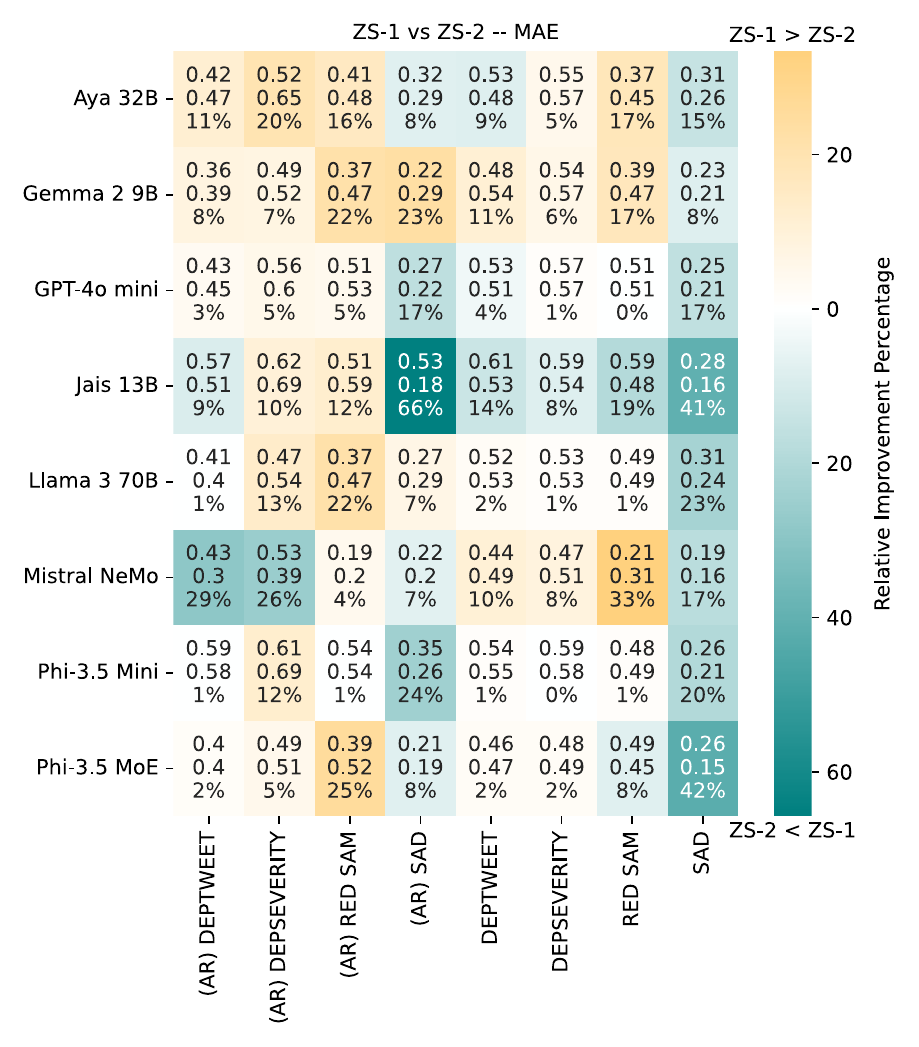}  %
        \caption{Comparison of model performance on MAE for prompts ZS-1 and ZS-2 across severity datasets, showing improvement percentage from the less performant prompt.}
    \label{fig:zs1vszs2_mae_all}      
\end{figure}

Continuing the prompt analysis for MAE, a heatmap is provided to compare performance across model-dataset pairs. In this heatmap, the first value represents the MAE for ZS-1, the second value is the MAE for ZS-2, and the third value is the relative improvement percentage, calculated as: 
\[
\text{Relative Improvement Percentage}_\text{MAE} = \left(1 - \frac{\text{Lower MAE}}{\text{Higher MAE}}\right) \times 100
\]
Here, the equation is different compared to the BA equation since lower MAE indicates better performance. The relative improvement percentage is used to determine the coloring of the heatmap shown in Fig. \ref{fig:zs1vszs2_mae_all}.

To compare the performance of prompts ZS-1 and ZS-2, a random fluctuation threshold of 5\% was defined. This threshold helps differentiate meaningful improvements from minor variations that could occur due to randomness or model variability. Specifically, any performance improvement smaller than 5\% was not considered a win for either prompt and instead categorized as a tie, as such small differences wouldn't be significant enough to be meaningful in judging prompt improvements.

Using this threshold, the analysis found that ZS-1 and ZS-2 each achieved \textbf{22 wins}, while \textbf{20 cases} were classified as ties. This suggests that, at first glance, both prompts generalize similarly across the tasks. Similarly, When considering the average relative improvement percentages, a mere \textbf{1.3\% advantage} was observed in favor of ZS-2. This indicates that prompt choice wasn't as impactful on this task.

The results presented in Table \ref{tab:model_prompt_sensitivity} highlight several noteworthy patterns regarding model sensitivity to prompt variations. Across all languages and tasks, Phi-3.5 MoE and Phi-3.5 Mini demonstrate the lowest levels of sensitivity, with average BA differences of only about 3–4. In contrast, Llama 3 70B and Aya 32B exhibit higher volatility, often exceeding 6 in average performance differences. These findings indicate that certain models are inherently more robust against prompt-induced fluctuations.

Moreover, the observed patterns remain consistent across different language configurations, including Arabic-only (AR), English-only (EN), Mixed (English Post + Arabic Prompt), and All (ALL). Models that exhibit stability in one language setting tend to maintain their resilience across others. Phi-3.5 MoE, for instance, is consistently among the least sensitive models regardless of the language configuration, showing only 3.2 BA variation in Arabic-only settings while retaining similarly low levels of fluctuation in EN, Mixed, and ALL experiments.

The type of classification task also plays a key role in model sensitivity. Multiclass classification tasks appear to provoke the most pronounced fluctuations. Llama 3 70B, for example, experiences a particularly high (34.5) variation for multi-class tasks in Arabic-only experiments, which stands in stark contrast to its more modest 4.7 variation for binary tasks. In comparison, Phi-3.5 MoE remains substantially more stable, with only a 4.2 difference reported in Arabic-only multi-class classification. These differences persist even when the complexity increases, as observed in the mixed-language experiments. Variation on MAE metric was negligible and thus it's not discussed.

\begin{table}[htbp]
\centering
\caption{Sensitivity of model BA performance to changes in prompt templates, reported across Arabic-only, English-only, and Mixed experiments (English Post + Arabic Prompt). Results are further categorized by task type.}
\label{tab:model_prompt_sensitivity}
\resizebox{\textwidth}{!}{
\begin{tabular}{l|rrrr|rrr|rrr|rrrr}
    \toprule
     Language & \multicolumn{4}{c|}{AR} & \multicolumn{3}{c|}{EN} & \multicolumn{3}{c|}{Mixed} & \multicolumn{4}{c}{ALL} \\
    DatasetType & ALL & Binary & Multi-Class & Severity & ALL & Binary & Severity & ALL & Binary & Multi-Class & ALL & Binary & Multi-Class & Severity \\
    Model &  &  &  &  &  &  &  &  &  &  &  &  &   &  \\
    \midrule
    AVG & 5.4 & 3.6 & 15.7 & 3.4 & 3.5 & 3.6 & 3.8 & 14.8 & 4.1 & 22.0 & 6.1 & 3.6 & 18.9 & 3.6 \\
    Llama 3 70B & 9.8 & 4.7 & 34.5 & 5.8 & 2.8 & 1.9 & 4.5 & 31.1 & 1.9 & 50.6 & 10.4 & 3.3 & 42.5 & 5.1 \\
    Aya 32B & 6.5 & 4.0 & 21.9 & 2.7 & 3.9 & 3.6 & 5.2 & 11.2 & 1.3 & 17.8 & 6.3 & 3.5 & 19.8 & 4.0 \\
    GPT-4o mini & 5.9 & 1.8 & 25.0 & 3.0 & 3.2 & 3.9 & 2.3 & 13.2 & 0.8 & 21.4 & 6.0 & 2.5 & 23.2 & 2.6 \\
    Gemma 2 9B & 5.6 & 2.3 & 21.5 & 3.1 & 1.8 & 1.0 & 3.1 & 16.5 & 2.8 & 25.7 & 5.8 & 1.8 & 23.6 & 3.1 \\
    Mistral NeMo & 4.8 & 6.0 & 5.3 & 4.0 & 5.7 & 8.2 & 2.4 & 22.3 & 19.5 & 24.1 & 7.6 & 8.4 & 14.7 & 3.2 \\
    Jais 13B & 3.8 & 3.4 & 5.2 & 3.2 & 4.7 & 5.5 & 4.3 & 11.9 & 1.2 & 19.1 & 5.2 & 4.0 & 12.2 & 3.7 \\
    Phi-3.5 Mini & 3.7 & 3.4 & 8.3 & 1.9 & 2.1 & 2.7 & 1.1 & 7.7 & 2.6 & 11.0 & 3.7 & 3.0 & 9.7 & 1.5 \\
    Phi-3.5 MoE & 3.2 & 3.0 & 4.2 & 3.8 & 3.8 & 2.0 & 7.6 & 4.9 & 2.6 & 6.4 & 3.6 & 2.5 & 5.3 & 5.7 \\
    \bottomrule
    \end{tabular}
    
}
    
\end{table}


\subsection{General Analysis}
\label{sec:general_analysis}
This section explores two critical questions:

\textbf{Q1: Which datasets posed the greatest challenges for the models, and which were the easiest?}  
Understanding the difficulty of different datasets provides valuable insights for future research and practical applications. Harder datasets can help uncover specific limitations of LLMs, while easier datasets set benchmarks for expected performance. For psychiatric model deployment, this information aids in predicting how models might handle real-world data entries.

To evaluate dataset difficulty, the rows in Figures \ref{fig:zs1vszs2_ba_all} and \ref{fig:zs1vszs2_mae_all} were normalized using their best values (maximum for BA and minimum for MAE). This normalization allows for a dataset-level difficulty assessment independent of the overall model performance. Table \ref{tab:dataset_difficulty} summarizes the results by averaging this metric across all models to rank datasets for each prompt.

For ease of interpretation: A dataset scoring 1 indicates that all models achieved their best performance on it. For BA, scores lower than 1 indicate that the models performed that fraction of what they achieved on their respective easiest dataset. The closer the score is to 0, the more difficult the dataset is relative to the easiest dataset. For MAE, scores higher than 1 reflect worse performance compared to the easiest dataset. The greater the score exceeds 1, the more the models deviated from their optimal error rate, highlighting greater difficulty in severity estimation.

The findings reveal that DCAT was the easiest dataset for the BA metric. In contrast, multi-class and severity datasets posed the greatest challenges. Two key observations emerge:\\
1. The AR and EN SDCNL Depression datasets were significantly more difficult than other binary datasets.\\
2. The MEDMCQA datasets (Native English, Google Translated Arabic, and BiMediX Arabic) exhibited a striking performance gap of 43.6\%. This disparity is further analyzed in Section \ref{sec:medmcqa}.

\begin{table}[htbp]
\centering
\caption{This table ranks dataset difficulty for each prompt based on evaluations by the selected LLMs. The entries are sorted by the "AVG" column, representing the average performance across prompts. Higher values indicate easier datasets for BA (Balanced Accuracy), while lower values indicate easier datasets for MAE (Mean Absolute Error). Standard deviation values next to the ratios provide additional context but do not affect the sorting.}
\label{tab:dataset_difficulty}
\begin{tabular}{llllll}
\toprule
 &  & ZS-1 & ZS-2 & AVG & Best \\
Metric & Dataset &  &  &  &  \\
\midrule
\multirow[t]{35}{*}{BA} & DCAT & 0.984 ± 0.018 & 0.993 ± 0.011 & 0.989 ± 0.015 & 0.988 ± 0.014 \\
 & CAIRODEP & 0.887 ± 0.041 & 0.881 ± 0.034 & 0.884 ± 0.038 & 0.882 ± 0.04 \\
 & (BINARIZED) DEPTWEET & 0.886 ± 0.088 & 0.872 ± 0.074 & 0.879 ± 0.081 & 0.891 ± 0.078 \\
 & (BINARIZED) SAD & 0.85 ± 0.066 & 0.904 ± 0.029 & 0.877 ± 0.048 & 0.891 ± 0.049 \\
 & MEDMCQA & 0.846 ± 0.138 & 0.851 ± 0.133 & 0.848 ± 0.135 & 0.842 ± 0.14 \\
 & (AR-BINARIZED) SAD & 0.803 ± 0.08 & 0.837 ± 0.023 & 0.82 ± 0.052 & 0.846 ± 0.023 \\
 & (AR-BINARIZED) DEPTWEET & 0.806 ± 0.085 & 0.8 ± 0.062 & 0.803 ± 0.074 & 0.809 ± 0.069 \\
 & (BINARIZED) DEPSEVERITY & 0.794 ± 0.079 & 0.8 ± 0.045 & 0.797 ± 0.062 & 0.81 ± 0.057 \\
 & (AR-BINARIZED) DEPSEVERITY & 0.773 ± 0.056 & 0.764 ± 0.052 & 0.768 ± 0.054 & 0.773 ± 0.05 \\
 & (EN) DCAT & 0.755 ± 0.362 & 0.732 ± 0.398 & 0.743 ± 0.38 & 0.76 ± 0.364 \\
 & DREADDIT & 0.683 ± 0.057 & 0.73 ± 0.057 & 0.707 ± 0.057 & 0.721 ± 0.06 \\
 & SDCNL SUICIDE & 0.719 ± 0.077 & 0.686 ± 0.082 & 0.703 ± 0.079 & 0.71 ± 0.083 \\
 & (EN) CAIRODEP & 0.709 ± 0.319 & 0.684 ± 0.337 & 0.696 ± 0.328 & 0.711 ± 0.32 \\
 & (AR) DREADDIT & 0.684 ± 0.074 & 0.691 ± 0.072 & 0.688 ± 0.073 & 0.706 ± 0.077 \\
 & (AR) SDCNL SUICIDE & 0.704 ± 0.054 & 0.644 ± 0.049 & 0.674 ± 0.051 & 0.694 ± 0.054 \\
 & (BINARIZED) RED SAM & 0.645 ± 0.051 & 0.639 ± 0.04 & 0.642 ± 0.046 & 0.645 ± 0.048 \\
 & (AR-BINARIZED) RED SAM & 0.638 ± 0.061 & 0.623 ± 0.068 & 0.63 ± 0.064 & 0.629 ± 0.06 \\
 & (AR) SDCNL DEPRESSION & 0.558 ± 0.028 & 0.575 ± 0.066 & 0.566 ± 0.047 & 0.562 ± 0.042 \\
 & SDCNL DEPRESSION & 0.545 ± 0.031 & 0.556 ± 0.062 & 0.55 ± 0.046 & 0.547 ± 0.037 \\
 & ARADEPSU & 0.444 ± 0.209 & 0.616 ± 0.196 & 0.53 ± 0.202 & 0.605 ± 0.186 \\
 & (AR) MEDMCQA-GT & 0.51 ± 0.15 & 0.528 ± 0.151 & 0.519 ± 0.15 & 0.524 ± 0.151 \\
 & DEPTWEET & 0.471 ± 0.066 & 0.539 ± 0.064 & 0.505 ± 0.065 & 0.531 ± 0.062 \\
 & (EN) ARADEPSU & 0.375 ± 0.272 & 0.546 ± 0.322 & 0.461 ± 0.297 & 0.584 ± 0.253 \\
 & (AR) DEPTWEET & 0.408 ± 0.064 & 0.454 ± 0.067 & 0.431 ± 0.066 & 0.445 ± 0.064 \\
 & (AR) MEDMCQA-BiMediX & 0.407 ± 0.083 & 0.416 ± 0.103 & 0.412 ± 0.093 & 0.416 ± 0.093 \\
 & MDE & 0.371 ± 0.141 & 0.441 ± 0.109 & 0.406 ± 0.125 & 0.466 ± 0.067 \\
 & (AR) RED SAM & 0.403 ± 0.022 & 0.407 ± 0.057 & 0.405 ± 0.04 & 0.408 ± 0.031 \\
 & RED SAM & 0.399 ± 0.038 & 0.409 ± 0.054 & 0.404 ± 0.046 & 0.406 ± 0.034 \\
 & DEPSEVERITY & 0.369 ± 0.028 & 0.393 ± 0.038 & 0.381 ± 0.033 & 0.387 ± 0.037 \\
 & (EN) MCD & 0.222 ± 0.164 & 0.498 ± 0.225 & 0.36 ± 0.195 & 0.501 ± 0.204 \\
 & (AR) DEPSEVERITY & 0.345 ± 0.035 & 0.358 ± 0.037 & 0.351 ± 0.036 & 0.355 ± 0.038 \\
 & (EN) MDE & 0.285 ± 0.184 & 0.397 ± 0.178 & 0.341 ± 0.181 & 0.432 ± 0.116 \\
 & MCD & 0.203 ± 0.147 & 0.413 ± 0.193 & 0.308 ± 0.17 & 0.416 ± 0.187 \\
 & SAD & 0.214 ± 0.048 & 0.274 ± 0.06 & 0.244 ± 0.054 & 0.269 ± 0.059 \\
 & (AR) SAD & 0.17 ± 0.054 & 0.247 ± 0.064 & 0.209 ± 0.059 & 0.241 ± 0.059 \\
\cline{1-6}
\multirow[t]{8}{*}{MAE} & SAD & 1.049 ± 0.085 & 1.0 ± 0.0 & 1.025 ± 0.043 & 1.0 ± 0.0 \\
 & (AR) SAD & 1.186 ± 0.312 & 1.215 ± 0.104 & 1.201 ± 0.208 & 1.166 ± 0.096 \\
 & (AR) RED SAM & 1.645 ± 0.383 & 2.437 ± 0.784 & 2.041 ± 0.584 & 2.1 ± 0.666 \\
 & (AR) DEPTWEET & 1.835 ± 0.341 & 2.26 ± 0.535 & 2.047 ± 0.438 & 2.2 ± 0.577 \\
 & RED SAM & 1.778 ± 0.434 & 2.343 ± 0.453 & 2.06 ± 0.443 & 2.167 ± 0.634 \\
 & DEPTWEET & 2.099 ± 0.195 & 2.661 ± 0.494 & 2.38 ± 0.344 & 2.568 ± 0.47 \\
 & DEPSEVERITY & 2.194 ± 0.248 & 2.82 ± 0.441 & 2.507 ± 0.344 & 2.732 ± 0.431 \\
 & (AR) DEPSEVERITY & 2.199 ± 0.357 & 2.95 ± 0.658 & 2.575 ± 0.507 & 2.685 ± 0.63 \\
\cline{1-6}
\bottomrule
\end{tabular}
\end{table}

\textbf{Q2: What is the ranking of models based on performance?}  
To assess model performance, each column in the model-dataset result matrix was normalized by dividing it by the maximum value for that column. This approach highlighted the relative performance of each model, independent of dataset difficulty. The normalized results were then averaged across all datasets to generate Table \ref{tab:general_model_compairson}. This analysis was performed in two configurations:
1. For the Arabic-only split, as it is a primary focus of this paper.
2. For all datasets combined, represented by the "AR" and "ALL" columns in the table.

Additionally, the analysis was conducted for each prompt (ZS-1 and ZS-2) and a hypothetical "Best" prompt, which represents the optimal performance prompt for each model-dataset pair. Models are ranked based on their "AVG" column scores, reflecting their average performance across ZS-1 and ZS-2.

The results show that Phi-3.5 MoE excelled in both BA and MAE metrics, ranking first and second, respectively. An interesting anomaly was observed with Mistral NeMo: it achieved the best MAE performance by a wide margin but ranked as the second-worst model for BA. This inconsistency is noticeable in both the Arabic-only and combined dataset groups.

\begin{table}[htbp]
\centering
\caption{Model performance ranking across all experiments, including Arabic-only datasets (native and translated). Results are presented for ZS-1, ZS-2, and the "Best" prompt. The "AVG" column, which averages performance across ZS-1 and ZS-2, determines the overall ranking. Higher values in BA and lower values in MAE indicate better performance.}
\label{tab:general_model_compairson}
\begin{tabular}{ll|rrrr|rrrr}
\toprule
 & Dataset Group & 
 \multicolumn{4}{r}{AR} & \multicolumn{4}{r}{ALL} \\
 \cline{1-10}
 & Prompt & ZS-1 & ZS-2 & AVG & Best & ZS-1 & ZS-2 & AVG & Best \\
 \cline{1-10}
Metric & Model &  &  &  &  &  &  &  &  \\
\midrule
\multirow[t]{8}{*}{BA} & Phi-3.5 MoE & 0.886 & 0.948 & 0.917 & 0.943 & 0.919 & 0.916 & 0.918 & 0.924 \\
 & Gemma 2 9B & 0.914 & 0.900 & 0.907 & 0.905 & 0.862 & 0.903 & 0.883 & 0.910 \\
 & GPT-4o Mini & 0.840 & 0.876 & 0.858 & 0.884 & 0.831 & 0.893 & 0.862 & 0.892 \\
 & Llama 3 70B & 0.774 & 0.880 & 0.827 & 0.888 & 0.797 & 0.920 & 0.858 & 0.924 \\
 & Aya 32B & 0.790 & 0.860 & 0.825 & 0.859 & 0.816 & 0.879 & 0.848 & 0.881 \\
 & Phi-3.5 Mini & 0.807 & 0.779 & 0.793 & 0.801 & 0.844 & 0.817 & 0.831 & 0.837 \\
 & Mistral NeMo & 0.759 & 0.725 & 0.742 & 0.759 & 0.703 & 0.687 & 0.695 & 0.737 \\
 & Jais 13B & 0.706 & 0.700 & 0.703 & 0.721 & 0.648 & 0.632 & 0.640 & 0.669 \\
\cline{1-10}
\multirow[t]{8}{*}{MAE} & Mistral NeMo & 1.087 & 1.023 & 1.055 & 1.023 & 1.043 & 1.028 & 1.036 & 1.018 \\
 & Phi-3.5 MoE & 1.288 & 1.567 & 1.427 & 1.401 & 1.367 & 1.339 & 1.353 & 1.355 \\
 & Gemma 2 9B & 1.252 & 1.634 & 1.443 & 1.386 & 1.291 & 1.470 & 1.380 & 1.382 \\
 & Llama 3 70B & 1.331 & 1.658 & 1.494 & 1.476 & 1.451 & 1.503 & 1.477 & 1.520 \\
 & Aya 32B & 1.480 & 1.810 & 1.645 & 1.609 & 1.464 & 1.582 & 1.523 & 1.531 \\
 & GPT-4o Mini & 1.582 & 1.718 & 1.650 & 1.681 & 1.562 & 1.517 & 1.540 & 1.613 \\
 & Phi-3.5 Mini & 1.851 & 1.955 & 1.903 & 1.927 & 1.701 & 1.645 & 1.673 & 1.741 \\
 & Jais 13B & 2.029 & 1.847 & 1.938 & 1.736 & 1.888 & 1.533 & 1.711 & 1.591 \\
\cline{1-10}
\bottomrule
\end{tabular}
\end{table}

\subsection{Language Analysis}
\subsubsection{English Native VS Arabic Translated Data}
\label{sec:english_to_arabic_analysis}
This section explores how performance differs between datasets in their original English versions and their English-to-Arabic translations (produced using Google Translate). The evaluation focuses on the best score (BA/MAE) achieved with either prompt (ZS-1 or ZS-2) to isolate the impact of language independently of prompt choice.

\begin{landscape}
    \begin{figure}[htbp]
    \centering
    \includegraphics[width=1.4\textwidth]{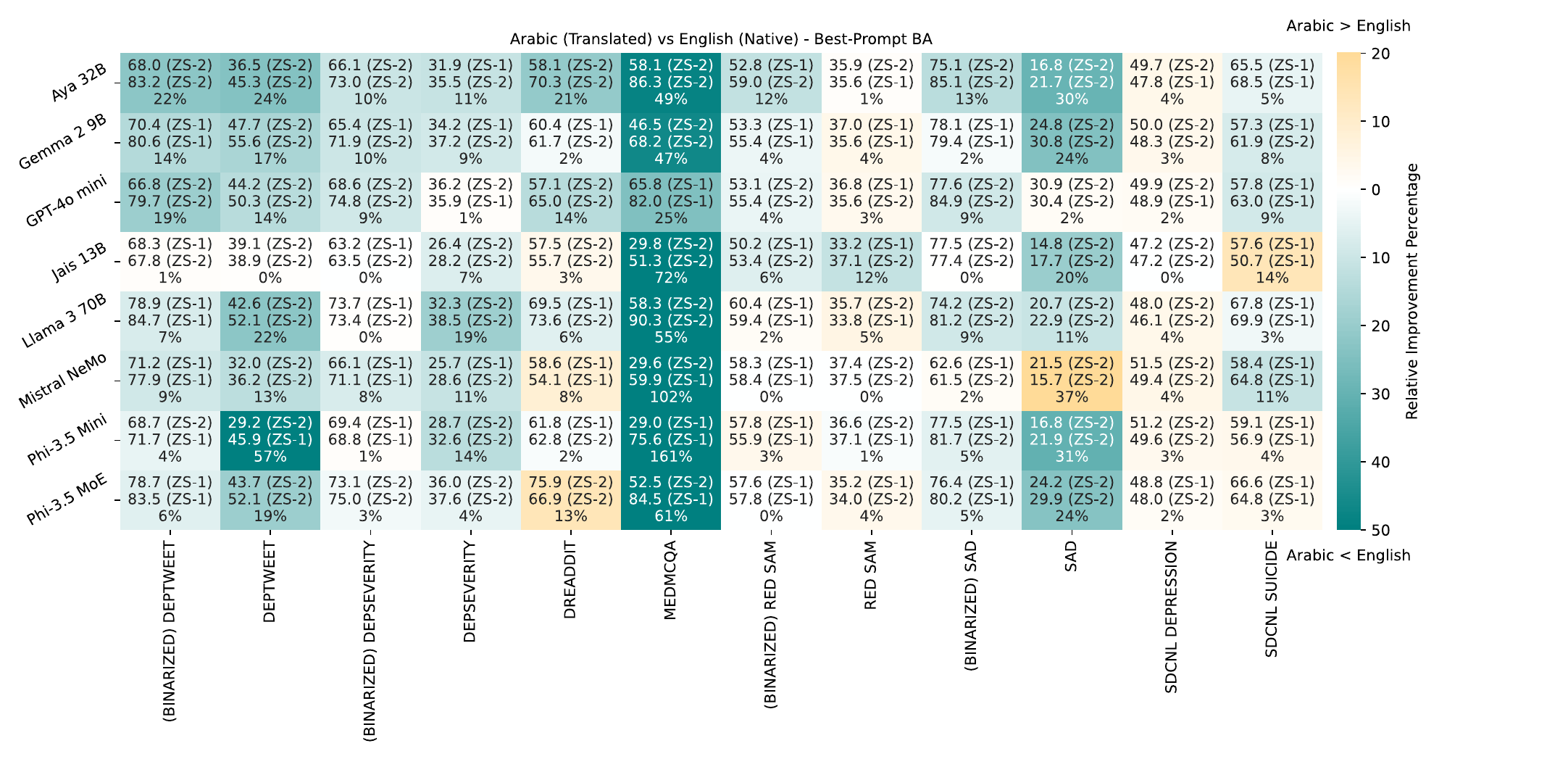}  %
    \caption{Comparison of BA for datasets in their original English versions versus their English-to-Arabic translations. Results are based on the best-performing prompt (ZS-1 or ZS-2), isolating the impact of language on performance.}
    \label{fig:tar_en_best_ba}          
\end{figure}
\end{landscape}

\begin{figure}[htbp]
    \centering
    \includegraphics[width=0.8\textwidth]{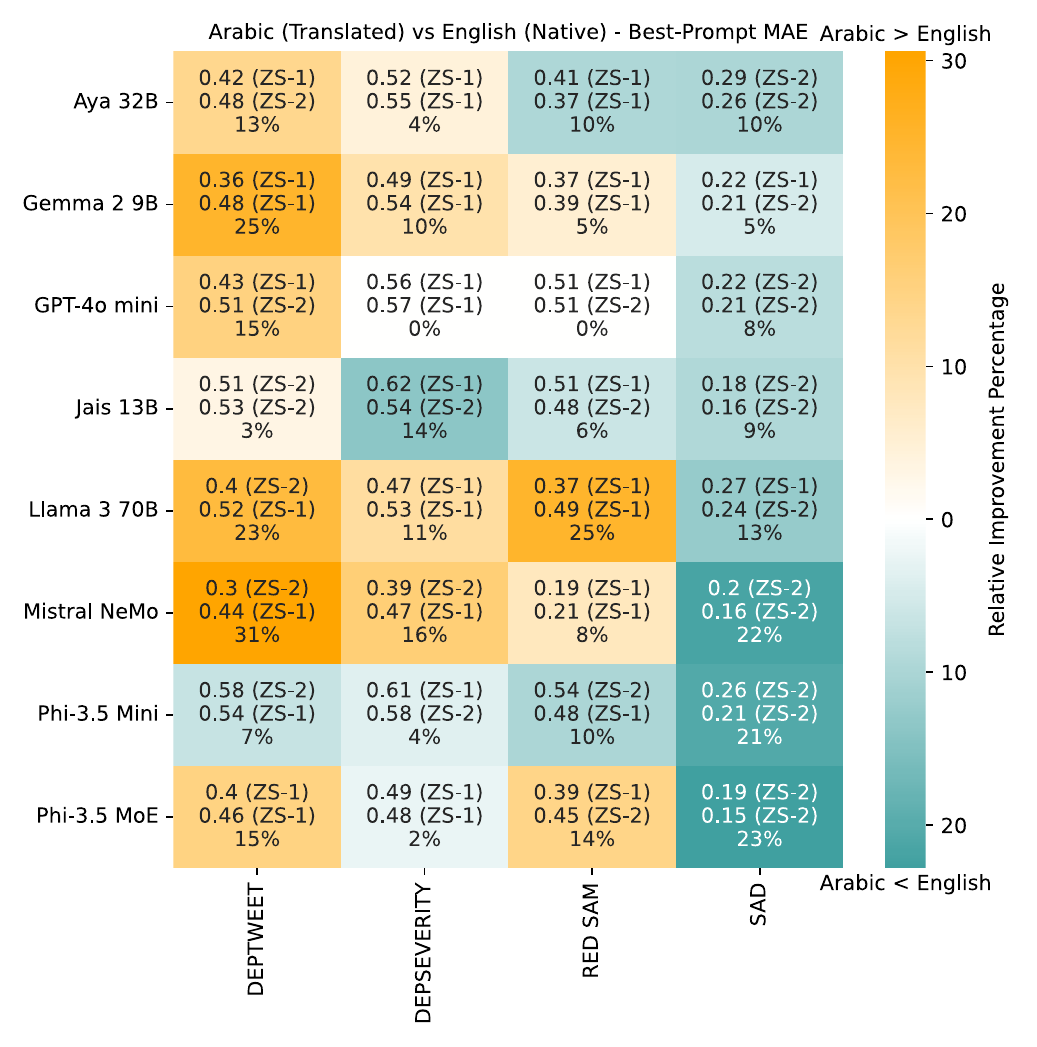}  %
    \caption{Comparison of MAE for datasets in their original English versions versus their English-to-Arabic translations. Results are based on the best-performing prompt (ZS-1 or ZS-2), isolating the impact of language on performance.}
    \label{fig:tar_en_best_mae}          
\end{figure}

Using the best-performing prompt for this comparison ensures that differences in performance are attributed to language rather than variations in prompt effectiveness on each language. Figure \ref{fig:tar_en_best_ba} presents the results for the balanced accuracy metric, while Figure \ref{fig:tar_en_best_mae} shows the results for the MAE metric in the severity datasets.

\textbf{Balanced Accuracy:}  
The heatmap illustrates the differences in balanced accuracy (BA) between the original English datasets and their Arabic translations. Each cell contains three values: the BA for the Arabic translation (top), the BA for the English version (middle), and the relative improvement percentage (bottom). The color of the cell indicates the direction of improvement: teal represents an advantage for the English version, while orange highlights cases where the Arabic version performed better. For instance, on the DREADDIT dataset, Phi-3.5 MoE performed 6\% better on the Arabic translation compared to the English version. In this specific example, ZS-2 was the best-performing prompt for both configurations. The prompts were translated one-to-one to match the dataset language.

The heatmap reveals that the English configuration outperformed the Arabic configuration in the majority of model-dataset pairs. A 5\% random fluctuation threshold was applied to distinguish meaningful differences from noise. Out of all runs, the Arabic version won only 5 times, compared to 51 wins for the English configuration, with 40 ties. These results align with expectations, as most models are trained predominantly on English datasets, and translation errors likely impact Arabic performance. Nonetheless, a few cases showed significant gains with the Arabic translation.

One notable anomaly is the MEDMCQA dataset, where the English version consistently outperformed the Arabic translation by more than 45\% across all models, with Phi-3.5 Mini showing a performance gap of 150\%. This disparity is explored in greater detail in Section \ref{sec:medmcqa}. Excluding this dataset, the average performance advantage of English over Arabic configurations is modest, at just 6\%.

\textbf{Mean Absolute Error}  
Figure \ref{fig:tar_en_best_mae} provides a similar analysis, focusing on the Mean Absolute Error (MAE), with the relative improvement percentage, displayed at the bottom of each cell, determining the heatmap's color scheme.

Unlike the balanced accuracy results, the figure suggests a different pattern for severity datasets. In several experiments, the Arabic version demonstrated a clear advantage over the English version, most notably on the DEPTWEET dataset. However, the SAD dataset consistently performed better when presented in its native language, regardless of the model used. Overall, the results showed 13 wins for Arabic, 12 for English, and 7 ties (Excluding experiments with less than 5\% performance difference). On average, the Arabic configuration had a mere advantage of 1.8\%, suggesting that language differences doesn't affect performance on this task task.

The heatmap lacks distinct vertical or horizontal trends, making it unclear whether performance differences stem from the models' language training or errors introduced during translation. To investigate this further, an additional language analysis experiment was conducted, as detailed in Section \ref{sec:native-arabic-vs-enlgish-translated}.

\subsubsection{Language Effect on MEDMCQA}
\label{sec:medmcqa}
The MEDMCQA dataset displayed the largest difference in balanced accuracy (BA) across models, as highlighted by the most intense teal color in the heatmap. On average, the native English configuration outperformed the Arabic version by 69\%, a striking gap compared to the second-largest difference of 21\% observed in other datasets. This discrepancy is unexpected and cannot be fully explained by the models’ generally reduced performance on Arabic datasets.

To investigate whether this degradation was caused by translation errors, an additional experiment was conducted using a version of MEDMCQA translated by the BiMediX \cite{bimedix} research paper. Unlike Google Translate, the BiMediX dataset employs a refined, semi-humanly supervised translation process, offering higher translation quality and consistency.

Table \ref{tab:medmcqa_comparison} compares the performance of the native dataset, the Google Translated version, and the BiMediX version, grouped by prompt and evaluated across all models. The results reveal that the Google Translated version generally outperformed the BiMediX translation, with differences as high as 16.7 for GPT-4o Mini on ZS-1. However, the native English dataset consistently achieved the highest accuracy, outperforming the Google Translated version by a factor of 1.2x for all prompts and models.

These findings suggest that language, translation quality and style significantly influence performance, with the native dataset’s substantial advantage hinting at potential training data leakage. 


\begin{table}[htbp]
\centering
\caption{Performance comparison of models and prompts on various MEDMCQA versions. The table is sorted by the "Bi - GT" column, which reflects the performance difference between the Arabic BiMediX and Arabic Google Translate (GTrans) versions. The "Native/GT" column shows the improvement factor when moving from the GTrans version to the native English dataset.}
\label{tab:medmcqa_comparison}
\begin{tabular}{llrrrrr}
\hline
 Prompt & Model  & English & Arabic BiMedix & Arabic GTrans & Bi - GT & Native / GT \\
\hline 
\multirow[t]{8}{*}{ZS-1} & Jais 13B & 50.6 & 26.7 & 26.9 & -0.2 & 1.9 \\
 & Mistral NeMo & 58.7 & 26.6 & 29.5 & -2.9 & 2.0 \\
 & Phi-3.5 Mini & 75.6 & 27.1 & 30.1 & -3.0 & 2.5 \\
 & Gemma 2 9B & 66.5 & 34.3 & 44.2 & -9.9 & 1.5 \\
 & Llama 3 70B & 87.9 & 42.8 & 53.2 & -10.4 & 1.6 \\
 & Phi-3.5 MoE & 84.3 & 38.6 & 52.2 & -13.6 & 1.6 \\
 & Aya 32B & 85.6 & 40.2 & 56.8 & -16.6 & 1.5 \\
 & GPT-4o Mini & 81.9 & 48.8 & 65.5 & -16.7 & 1.2 \\
\cline{1-7}
\multirow[t]{8}{*}{ZS-2} & Phi-3.5 Mini & 73.5 & 28.7 & 27.6 & 1.1 & 2.7 \\
 & Mistral NeMo & 58.1 & 26.5 & 30.1 & -3.6 & 1.9 \\
 & Jais 13B & 51.3 & 21.4 & 30.8 & -9.4 & 1.7 \\
 & Gemma 2 9B & 68.4 & 34.9 & 46.9 & -12.0 & 1.5 \\
 & Phi-3.5 MoE & 83.6 & 40.6 & 52.8 & -12.2 & 1.6 \\
 & Llama 3 70B & 90.2 & 44.3 & 58.1 & -13.8 & 1.6 \\
 & GPT-4o Mini & 80.7 & 50.4 & 64.8 & -14.4 & 1.2 \\
 & Aya 32B & 86.1 & 43.6 & 58.1 & -14.5 & 1.5 \\
\cline{1-7}
\multirow[t]{8}{*}{Best-Prompt} & Phi-3.5 Mini & 75.6 & 28.7 & 30.1 & -1.4 & 2.5 \\
 & Mistral NeMo & 58.7 & 26.6 & 30.1 & -3.5 & 1.9 \\
 & Jais 13B & 51.3 & 26.7 & 30.8 & -4.1 & 1.7 \\
 & Gemma 2 9B & 68.4 & 34.9 & 46.9 & -12.0 & 1.5 \\
 & Phi-3.5 MoE & 84.3 & 40.6 & 52.8 & -12.2 & 1.6 \\
 & Llama 3 70B & 90.2 & 44.3 & 58.1 & -13.8 & 1.6 \\
 & Aya 32B & 86.1 & 43.6 & 58.1 & -14.5 & 1.5 \\
 & GPT-4o Mini & 81.9 & 50.4 & 65.5 & -15.1 & 1.2 \\
\cline{1-7}
\bottomrule
\end{tabular}
\end{table}

\subsubsection{Arabic Native VS English Translated}
\label{sec:native-arabic-vs-enlgish-translated}
In this experiment, two key changes were made compared to the earlier English-to-Arabic analysis. First, the direction of translation was reversed: the Arabic native datasets were translated into English using Google Translate. While the translation direction differed, the method (Google Translate) remained the same as in the previous experiment. Second, the prompt template remained in Arabic, identical to that used for the Arabic datasets, with only the dataset posts being translated into English. This setup was designed to isolate the effect of language on performance while addressing the following observations:
1. Models demonstrated an understanding of Arabic prompts in the invalid response analysis.
2. Modifying the prompt, even randomly, could influence performance.
3. To fairly evaluate language effects, it was most logical to keep the prompt instructions in Arabic, regardless of the post language.

\begin{figure}[htbp]
    \centering
\includegraphics[width=0.85\textwidth]{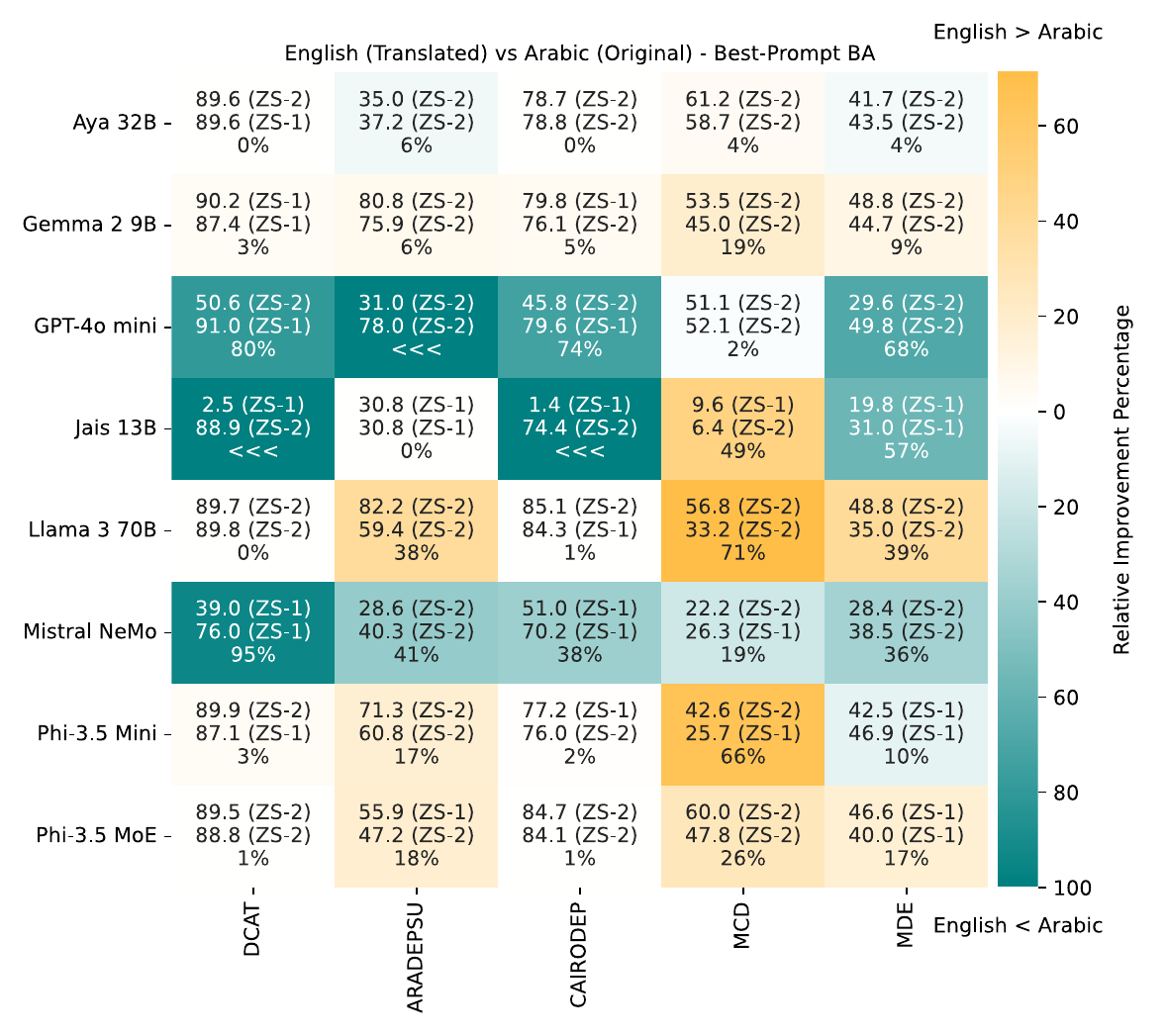}  %
    \caption{Comparison of BA performance between translated English and original Arabic versions of the same dataset. Results are based on the highest-scoring prompt for each configuration. The symbol \texttt{<}{\texttt{<}}{\texttt{<}} indicate that the improvement is greater than 100\%, which avoids displaying excessively large numbers when the worse prompt’s performance approaches zero.}
    \label{fig:ten_ar_best_ba}          
\end{figure}

As with the previous analysis, results are based on the best-scoring prompt for each configuration. Figure \ref{fig:ten_ar_best_ba} follows the same heatmap format as earlier figures.

The figure highlights performance differences across models. GPT-4o Mini, Jais 13B, and Mistral NeMo performed better on the original Arabic datasets, while Llama 3 70B and Phi-3.5 MoE achieved better results on the translated versions. Aya 32B was largely unaffected by the language change, with no notable gains in either direction for any dataset. In total, there were 12 wins for the English configuration, 14 for the Arabic configuration, and 14 ties. The average difference in performance between the two configurations is 6.3 in direction of Arabic. This variation was driven by an average increase of 17.3 on binary task as multi-class performance difference was negligible. Taking only the samples where both prompts led to non-invalid parses, the average increase reduced to 8.7, meaning that the difference was partially due to improved instruction following in the case of non-mixed Arabic language.


\subsection{Few-Shot Experiment}
\label{sec:fewshot}
This section describes a few-shot experiment conducted on two models: Phi-3.5 MoE and GPT-4o Mini. These models were selected as the best-performing open-source and proprietary options, respectively. The experiment used Prompt-2 exclusively, as it demonstrated better performance in prior analyses. Guiding examples were appended to the instruction template, positioned before the test post for which the model's prediction was required. 

The guiding examples were chosen based on the following criteria:\\
- For datasets with more than 1,000 samples, examples were randomly sampled from outside the test dataset.\\
- If fewer than 1,000 samples were available, the selected examples were omitted from the dataset.\\ 

The number of guiding examples corresponded to the number of classes in the dataset, with one random example drawn from each class.

\begin{figure}[htb]
\begin{mdframed}[backgroundcolor=gray!10, linecolor=gray!10]
    \centering
    \textbf{Example Arabic FS-2 Prompt (Prompt 1 for Binary Depression Detection)}\\[2mm]
        \includegraphics[width=1\linewidth]{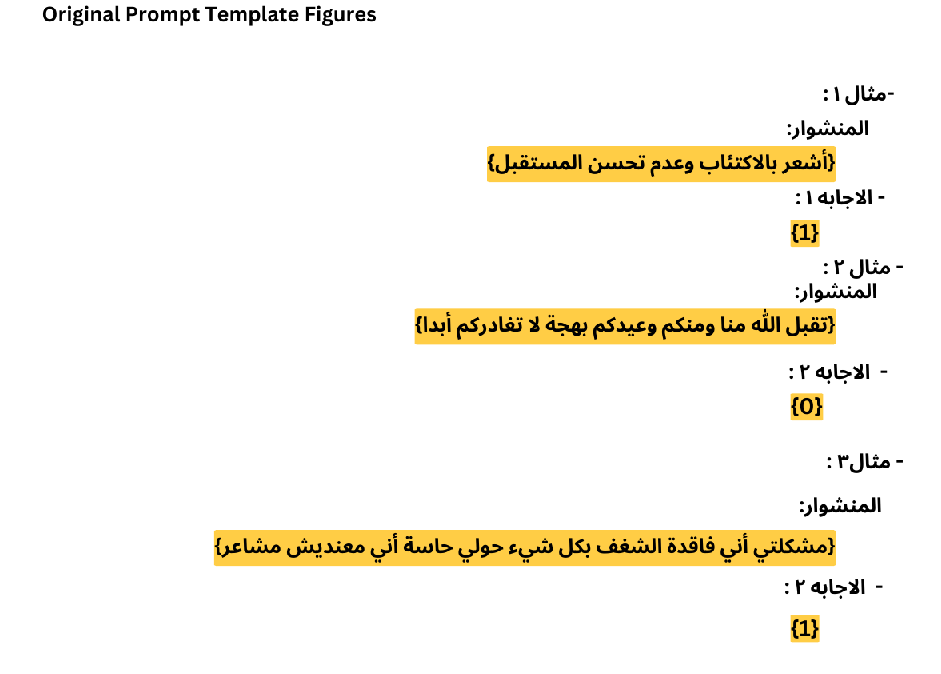}

    \caption{An example of a few-shot (FS-2) prompt for binary depression detection. The prompt includes multiple examples to guide the model on how to classify posts with corresponding answers.}
    \label{fig:example-arabic}
\end{mdframed}
\end{figure}

\begin{figure}[htb]
\begin{mdframed}[backgroundcolor=gray!8, linecolor=gray!8]
    \centering
    \textbf{Example English Prompt (Prompt 1 for Binary Depression Detection)}\\[2mm]
    \begin{tikzpicture}
        \node[fill=white!5, draw=gray, rounded corners, inner sep=10pt, text width=0.95\textwidth] (box1) {
            \textbf{ Original Prompt Template Figures~\ref{fig:fs-p_1}} \\[2mm]
            \textbf{- Example 1:}\\
            \hspace*{5mm} Post: \\
            \quad \quad \colorbox[HTML]{EFB61F}{ {\{I feel depressed and the future is not improving\}}}\\
            \hspace*{5mm} Answer: \\
            \quad \quad  \colorbox[HTML]{EFB61F}{{\{1\}}}\\[2mm] 
            \textbf{- Example 2:}\\
            \hspace*{5mm} Post: \\
            \quad \quad  \colorbox[HTML]{EFB61F}{ {\{May Allah accept from us and from you and may your Eid be a joy that never leaves you\}}}\\
            \hspace*{5mm} Answer: \\
            \quad \quad \colorbox[HTML]{EFB61F}{{\{0\}}}\\[2mm]  
            \textbf{- Example 3:}\\
            \hspace*{5mm} Post: \\
            \quad \quad \colorbox[HTML]{EFB61F}{{\{ My problem is that I have lost passion for everything around me. I feel like I have no feelings\}}}\\
            \hspace*{5mm} Answer: \\
            \quad \quad \colorbox[HTML]{EFB61F}{{\{1\}}}\\[2mm]  
            \textbf{Post to be analyzed: \\}
            \hspace*{5mm} \colorbox[HTML]{EFB61F}{{ \textbf{\{XXXXXX\}}}}
        };
    \end{tikzpicture}
    \caption{An example of a few-shot (FS-2) prompt for binary depression detection. The prompt includes multiple examples to guide the model on how to classify posts with corresponding answers.}
    \label{fig:fS_Example}
\end{mdframed}
\end{figure}

\begin{figure}[htbp]
    \centering
    \includegraphics[width=1\textwidth]
    {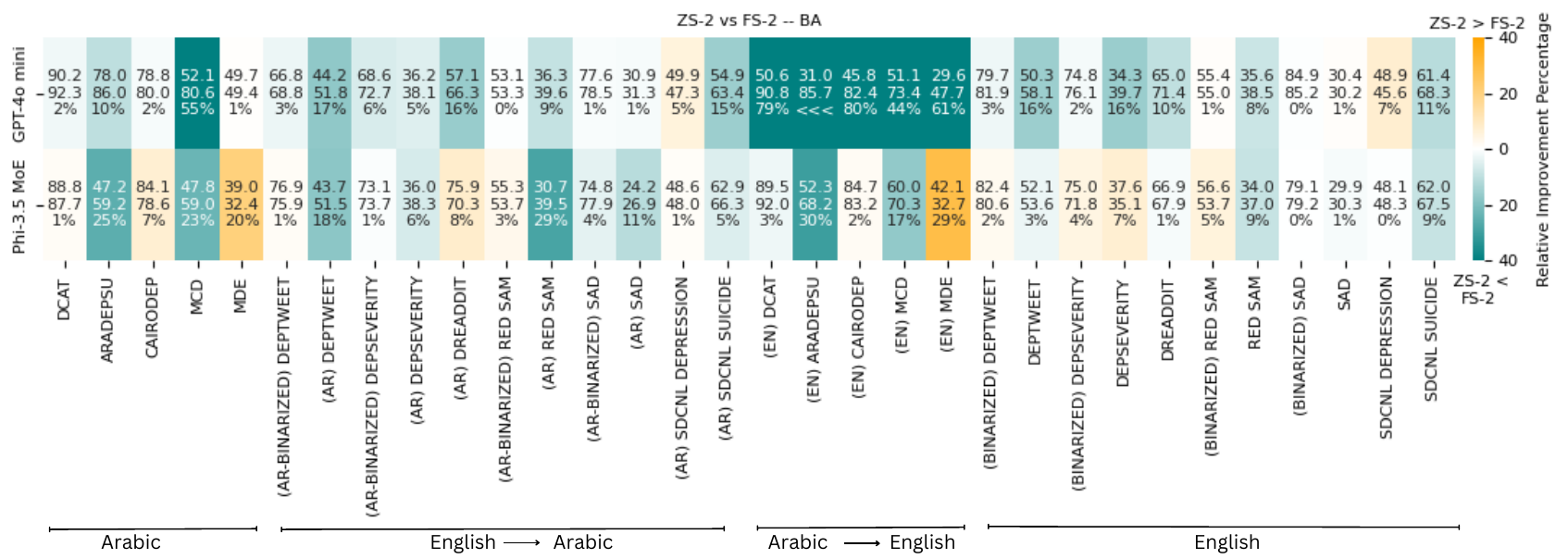}
    \caption{Comparison of BA performance for each model and dataset when switching from ZS-2 to FS-2. Each cell displays the BA for ZS-2 (top), FS-2 (middle), and the percentage improvement (bottom).}
    \label{fig:fs_zs_ba_comparison}
\end{figure}

\begin{figure}[htbp]
    \centering
\includegraphics[width=0.75\textwidth]
    {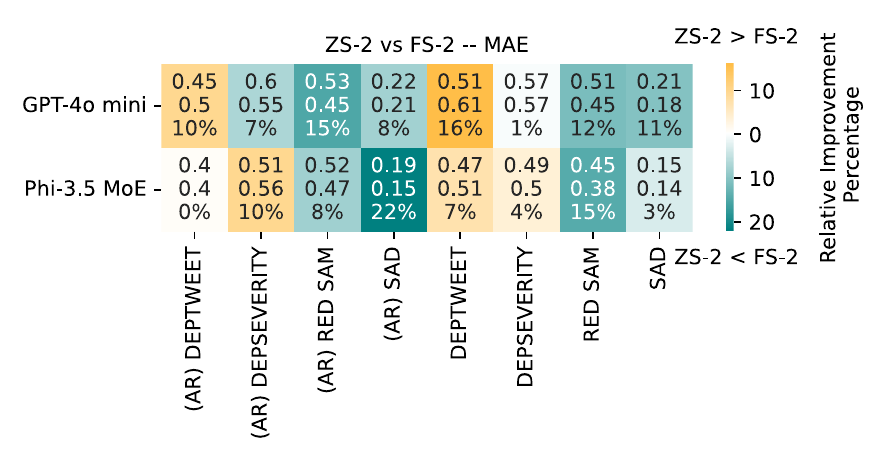}
    \caption{Comparison of MAE performance for each model and dataset when switching from ZS-2 to FS-2. Each cell displays the MAE for ZS-2 (top), FS-2 (middle), and the percentage improvement (bottom).}
    \label{fig:fs_zs_mae_comparison}
\end{figure}

Figures \ref{fig:fs_zs_ba_comparison} and \ref{fig:fs_zs_mae_comparison} summarize the results for BA and MAE metrics, respectively. The inclusion of guiding examples significantly enhanced GPT-4o Mini's performance, with a mean improvement of 20\% for FS-2 compared to ZS-2. The most substantial gains were observed on multi-class datasets, where performance improved by 57.6\%. Phi-3.5 MoE also benefitted from the additional examples, showing average improvements of 3.3\% and 7.9\% on multi-class datasets. Both models achieved an average improvement of 8.8\% on severity datasets. For binary datasets, GPT-4o Mini improved by 12\%, while Phi-3.5 MoE experienced a negligible performance loss of less than 1\%.

The impact of few-shot prompting on Arabic-only datasets was less significant for GPT-4o Mini, with improvements reduced to 21.4\% and 4.5\% for multi-class and binary datasets, respectively. For Phi-3.5 MoE, the effects were generally minimal, except for severity datasets, where improvements reached 16\%. These results are presented in Table \ref{tab:fs_effect_table}.

Overall, the results demonstrate that few-shot prompting is an effective method for boosting model performance, particularly for complex multi-class tasks.

\begin{table}[]
\centering
\caption{Average improvement in BA performance (calculated across datasets) when transitioning from ZS-2 to FS-2. Results are presented for all datasets and Arabic-only datasets, with additional sub-grouping by task type.}
\label{tab:fs_effect_table}
\begin{tabular}{l|rrrr|rrrr}
\toprule
 & \multicolumn{4}{c}{ALL} & \multicolumn{4}{c}{AR} \\
 & Multi & Binary & Severity & All & Multi & Binary & Severity & All \\
\midrule
GPT-4o Mini & 57.60 & 12.10 & 8.95 & 19.84 & 21.45 & 4.50 & 8.21 & 8.60 \\
Phi-3.5 MoE & 7.88 & -0.64 & 8.75 & 3.30 & 9.49 & -1.27 & 16.01 & 5.07 \\
\bottomrule
\end{tabular}
\end{table}

For MAE, the improvements were more modest, averaging 3.5\% for both models. This figure increased slightly to 5.15\% when considering the Arabic-only dataset group.

\section{Conclusion and Future Work}
\label{sec:section 5}
This study revealed key factors influencing the performance of large language models (LLMs) on psychiatric diagnostic tasks. Prompt design significantly impacted results with even slight changes altering instruction adherence and causing parsing failures, particularly in ZS-1, as ZS-2 improved over ZS-1 by 14.5 BA on multi-class datasets. Model choice was critical: Phi-3.5 MoE excelled in BA on Arabic datasets, followed by Gemma 2 9B, while Mistral NeMo dominated MAE tasks, outperforming others by a wide margin.

Dataset difficulty varied substantially, with DCAT performing best across models, while datasets like SDCNL Depression approached random guessing. This highlights the need to investigate whether such disparities stem from labeling quality or inherent diagnostic challenges. 

Language analysis indicated that the effect of language on performance was modest, as English configuration improved results of Arabic by an average of 6\% on BA. Translating Arabic datasets into English while maintaining Arabic prompt instructions suggested that performance differences were primarily driven by translation quality and style, as Arabic achieved a higher performance, reaching a difference of 17.3 on BA task on binary datasets.

Few-shot prompting improved outcomes, particularly for GPT-4o Mini, with a 20\% BA gain on average and 57.6\% improvement on multi-class tasks. Phi-3.5 MoE also benefitted, albeit more modestly. These findings underscore the importance of tailoring prompts, leveraging few-shot techniques, and addressing language biases to optimize LLM performance. Future work should expand multilingual datasets and explore more refined prompting strategies to unlock the full potential of LLMs in psychiatric diagnostics.

One limitation of this study was querying LLMs for a single-word diagnosis. Future research could explore prompting models to provide reasoning after diagnosis, which would offer insights into their interpretability and true understanding of tasks. Additionally, experiments involving reasoning before diagnosis, such as chain-of-thought prompting, could leverage the causal reasoning abilities of LLMs and potentially improve diagnostic accuracy.

This study primarily focused on observing performance across a broad range of models and datasets. Future work could narrow the scope to investigate specific models or datasets of interest, allowing for deeper exploration of model behaviors and a more detailed understanding of certain patterns and anomalies.

As this study relied on machine-translated datasets, conclusions regarding language effects might also reflect translation quality. A follow-up study using expertly translated datasets could better isolate and evaluate pure language-induced performance changes, enabling more precise insights into cross-lingual capabilities.

Further research could examine the adaptability of LLMs to psychiatric diagnostics by testing techniques such as in-context learning or fine-tuning on small, domain-specific datasets. These methods could enhance model performance and their ability to generalize to unseen tasks with minimal additional training.

Additionally, deeper error analysis focusing on incorrect diagnoses could uncover patterns in model failures, guiding the development of robust prompting strategies or dataset enhancements to mitigate these issues. Robustness testing could also provide insights into how LLMs handle noisy or ambiguous inputs, further improving their reliability.

By addressing these directions, future research can build upon the findings of this study, advancing the development and deployment of LLMs in critical diagnostic tasks.




\section{Appendix}

\subsection{Dataset Description}  
\label{sec:dataset-stats}
Figure \ref{tab:dataset-stats} summarizes the class distributions for all datasets used in this study. These distributions are essential for understanding sampling balance and evaluating how fair-random sampling might influence model performance. In addition to class distributions, sample length statistics—mean, standard deviation, minimum, and maximum word counts—are reported for the Arabic versions of each dataset. For datasets originally in Arabic, the statistics are calculated on the native versions, while for those translated from English, they are derived from the Google Translated versions.  

Notably, the statistics reveal that several datasets include extremely short samples, with minimum lengths as low as one word. This observation highlights potential concerns about data quality, which could affect both the models' understanding and their overall performance.  

\begin{table}[htbp!]
\begingroup
\renewcommand{\arraystretch}{1.1}
\label{tab:dataset-stats} %
\caption{Statistics of datasets included in the study. The table provides details on dataset names, class distributions, sample counts, and sample length statistics (mean, standard deviation (Std), minimum (Min), and maximum (Max) word counts). Length statistics are reported for the Arabic versions of the datasets, either native or translated from English.}

  \resizebox{\columnwidth}{!}{%
  \begin{tabular}{lll|cccc|}
  \cline{4-7}
   & & &
 \multicolumn{4}{c|}{Sample Length Statistic} \\ \hline
 \multicolumn{1}{l|}{Dataset} &
Class & Count & Mean & Std & Min & Max\\ \hline
\multicolumn{1}{l|}{DCAT} &
    \begin{tabular}[c]{@{}l@{}}- Depression\\ - Non Depression\end{tabular} &
    \begin{tabular}[c]{@{}l@{}}- 5,000\\ - 5,000\end{tabular} &
    10.8 &
    6.7 &
    1 &
    59\\ \hline
  \multicolumn{1}{l|}{MCD} &
    \begin{tabular}[c]{@{}l@{}}- Diminished ability to think or concentrate\\ - Psychomotor agitation or retardation\\ - Sleep disorder\\ - Weight disorder \\ - Low mood\\ - Loss of energy \\ - Feelings of worthlessness \\ - Suicidality \\ - Losing interest or pleasure in activities\end{tabular} &
    \begin{tabular}[c]{@{}l@{}}- 270\\ - 120\\ - 117\\ - 110\\ - 109\\ - 101\\ - 96\\ - 95\\ - 76\end{tabular} &
    7.3 &
    4.5 &
    1 &
    34\\ \hline
  \multicolumn{1}{l|}{ARADEPSU} &
    \begin{tabular}[c]{@{}l@{}}- Non-depression\\ - Depression Mood\\ - Depression With Suicidal Ideation\end{tabular} &
    \begin{tabular}[c]{@{}l@{}}- 3,166\\ - 1,355\\ - 531\end{tabular} &
    18.8 &
    38.3 &
    1 &
    1,432\\ \hline
  \multicolumn{1}{l|}{CAIRODEP} &
    \begin{tabular}[c]{@{}l@{}}- Depression\\ - Non Depression\end{tabular} &
    \begin{tabular}[c]{@{}l@{}}- 3,586\\ - 3,205\end{tabular} &
    86.8 &
    143.8 &
    1 &
    2,128\\ \hline
  \multicolumn{1}{l|}{AMI} &
    \begin{tabular}[c]{@{}l@{}}- Insomnia\\ - Depression\\ - Anxiety\\ - Bipolar\\ - Stress\end{tabular} &
    \begin{tabular}[c]{@{}l@{}}- 235\\ - 216\\ - 136\\ - 58\\ - 54\end{tabular} &
    13.2 &
    6.6 &
    2 &
    29\\ \hline
  \multicolumn{1}{l|}{MDE} &
    \begin{tabular}[c]{@{}l@{}}- Suicide\\ - Anxiety\\ - Depression\end{tabular} &
    \begin{tabular}[c]{@{}l@{}}- 600\\ - 597\\ - 600\end{tabular} &
    12.6 &
    7 &
    3 &
    82\\ \hline
  \multicolumn{1}{l|}{DEPSEVERITY} &
    \begin{tabular}[c]{@{}l@{}}Depression :\\ - Minimal \\ - Mild\\ - Moderate\\ - Severe\end{tabular} &
    \begin{tabular}[c]{@{}l@{}}\\- 2,587\\ - 394\\ - 290\\ - 282\end{tabular} &
    71.6 &
    27.2 &
    1 &
    262\\ \hline
  \multicolumn{1}{l|}{DEPTWEET} &
    \begin{tabular}[c]{@{}l@{}}Depression :\\ - Not Depressed\\ - Mild\\ - Moderate\\ - Severe\end{tabular} &
    \begin{tabular}[c]{@{}l@{}}\\- 32,400\\ - 5,242\\ - 1,809\\ - 740\end{tabular} &
    25.8 &
    13.6 &
    4 &
    62\\ \hline
  \multicolumn{1}{l|}{DREADDIT} &
    \begin{tabular}[c]{@{}l@{}}- Depression\\ - Non Depression\end{tabular} &
    \begin{tabular}[c]{@{}l@{}}- 369\\ - 346\end{tabular} &
    71.6 &
    26.8 &
    13 &
    229\\ \hline
  \multicolumn{1}{l|}{RED SAM} &
    \begin{tabular}[c]{@{}l@{}}Depression :\\ - Not Depressed\\ - Moderate\\ - Severe\end{tabular} &
    \begin{tabular}[c]{@{}l@{}}\\- 848\\ - 2,169\\ - 228\end{tabular} &
    138.7 &
    169.6 &
    1 &
    2,659\\ \hline
  \multicolumn{1}{l|}{SAD} &
    - Severity 0 to 10 &
    - 6,850 &
    10.2 &
    5 &
    1 &
    35\\ \hline
  \multicolumn{1}{l|}{SDCNL} &
    \begin{tabular}[c]{@{}l@{}}- Depression\\ - Suicide\end{tabular} &
    \begin{tabular}[c]{@{}l@{}}- 186\\ - 193\end{tabular} &
    139 &
    157.6 &
    7 &
    1,499\\
\hline
  \end{tabular}%
  }
  \endgroup
  \end{table}

\subsection{Preliminary Evaluation of Arabic Models}
\label{sec:arabic_models}

To identify high-performing Arabic-focused models for inclusion in the main evaluation, a preliminary experiment was conducted using randomly sampled data from two Arabic datasets: DCAT and CAIRODEP. This experiment assessed the performance of several Arabic large language models (LLMs): Aya, Jais, Silma, Noon, AceGPT, and ar-stablelm-2.  

The results of this evaluation are summarized in Table \ref{tab:arabic_models}. Aya and Jais emerged as the top-performing models, with Aya achieving the highest scores on both datasets, followed closely by Jais. These results highlight the superior performance of larger models like Aya 32B and Jais 13B, while smaller models, such as AceGPT (8B), achieved relatively lower scores.  

Quantization techniques, including 4-bit and 8-bit precision, were applied to some models to reduce computational requirements, which may have influenced performance. The inclusion of Aya and Jais in the main study is based on their consistent performance and computational feasibility. 

\begin{table}[ht]
\centering
\resizebox{\columnwidth}{!}{%
\begin{tabular}{c|c|c|c|c|c|c}
\hline
\textbf{Model} & \textbf{Size (Parameters)} & \textbf{Source} & \textbf{Huggingface Name}         & \textbf{DCAT} & \textbf{CAIRODEP} & \textbf{Quantization} \\ \hline
Aya            & 32B                        & Cohere          & c4ai-aya-expanse-32b               & 89                 & 81.7              & -                     \\
Jais           & 13B                        & Huggingface     & inceptionai/jais-family-13b-chat   & 88.9               & 74.4              & 8-bit                 \\
Silma          & 9B                         & Huggingface     & silma-ai/SILMA-9B-Instruct-v1.0    & 65.1               & 80.7              & 4-bit                 \\
Noon           & 7B                         & Huggingface     & Naseej/noon-7b                     & 55.4               & 59.2              & 8-bit                 \\
Acegpt         & 8B                         & Huggingface     & FreedomIntelligence/AceGPT-8B-chat & 47.2               & 57.5              & -                     \\ 
ar-stablelm-2         & 1.64B                         & Huggingface     & Fstabilityai/ar-stablelm-2-chat &           26.4    &         27.9      & -                    \\ \hline

\end{tabular}%
}
\caption{Preliminary evaluation results of Arabic LLMs on DCAT and CAIRODEP datasets.}
\label{tab:arabic_models}
\end{table}

\subsection{Response Length Analysis}
\label{sec:response_length_analysis}
Section \ref{sec:invalid_response_analysis} Analyzed the \% of invalid model responses, However this only indicated that a parsing failed, parsing could fail due to two reasons, the response was too wordy and thus contradicting keywords were found, or the response didn't include any of the target keywords.

\begin{figure}[htbp]
    \centering
    \includegraphics[width=1\textwidth]{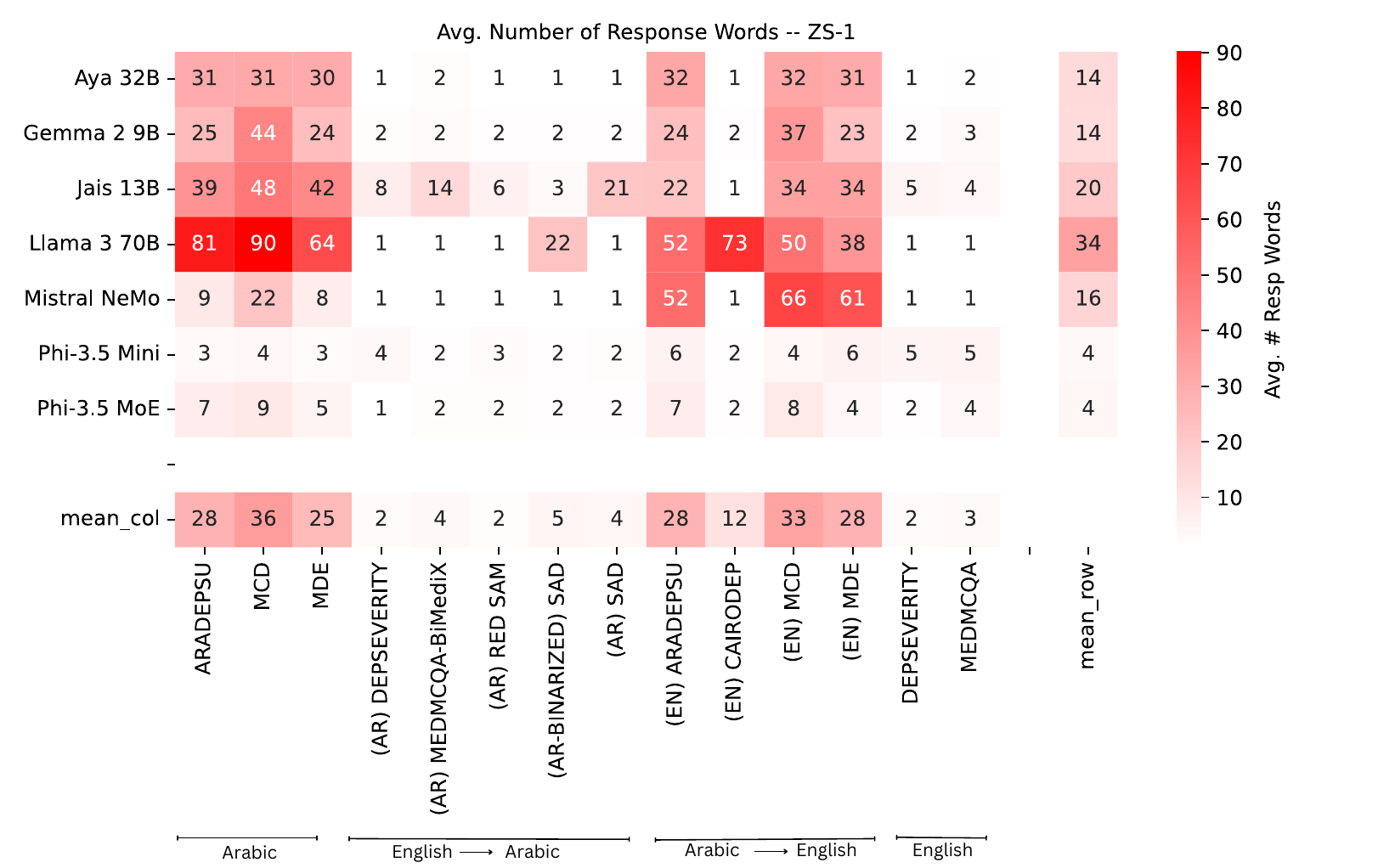}  %
    \caption{Average response length (words) for each model, dataset pair on ZS-1. Columns with a maximum value < 5 were omitted for clarity.}
    \label{fig:word_count_zs1_trunc}          
\end{figure}

\begin{figure}[htbp]
    \centering
    \includegraphics[width=1\textwidth]{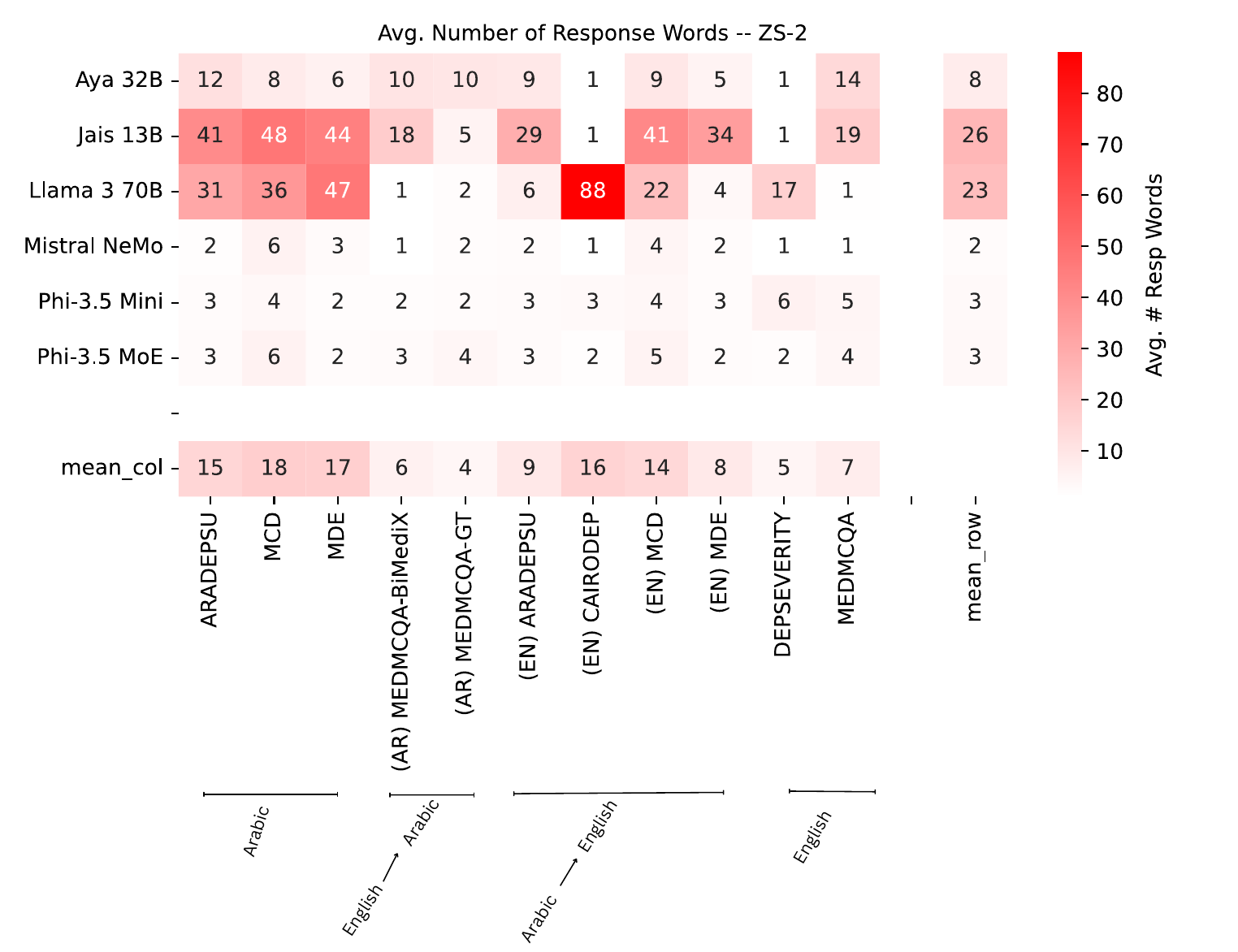}  %
    \caption{Average response length (words) for each model, dataset pair on ZS-2. Columns with a maximum value < 5 were omitted for clarity.}
    \label{fig:word_count_zs2_trunc}          
\end{figure}

This section provide an analysis of the average response length as shown in Figures \ref{fig:word_count_zs1_trunc} and \ref{fig:word_count_zs2_trunc}.Serving two purposes, discovering when the parse failed due to absence of a keyword (this is deducible if the response length is small but the invalid \% responses is high), and it also serve to rank the models on instruction following on dataset, prompt pairs.

\vspace{10cm}

\subsection{AMI Dataset Results}  

A severe abnormality was observed in the scores of all models on the AMI dataset, where none of the models achieved higher than random guessing (50\%) on any of the dataset-disorder splits, as shown in Table \ref{tab:AMI_table}. As a result, this dataset was omitted from the main results and discussion section.  

Further manual investigation into the dataset samples revealed widespread issues with faulty labeling, which likely contributed to the poor performance across all models. This inconsistency raises concerns about the dataset's reliability for evaluating model performance in psychiatric diagnosis tasks.

\label{sec:AMI}
\begin{table}[htbp]
    \centering
    \caption{BA performance of models on different disorders within the AMI dataset. The reported performance corresponds to the maximum score achieved between ZS-1 and ZS-2 prompts.}
    \begin{tabular}{llllll}
    \toprule
     & Anxiety & Bi-Polar & Depression & Insomia & Stress \\
    \midrule
    Gemma 2 9B & 42.6 (ZS-2) & 39.3 (ZS-2) & 30.7 (ZS-2) & 43.0 (ZS-2) & 45.2 (ZS-1) \\
    GPT-4o mini & 45.2 (ZS-1) & 43.7 (ZS-1) & 32.3 (ZS-1) & 46.5 (ZS-1) & 47.6 (ZS-1) \\
    Jais 13B & 38.8 (ZS-2) & 42.3 (ZS-2) & 31.6 (ZS-1) & 47.9 (ZS-1) & 42.9 (ZS-1) \\
    Llama 3 70B & 38.8 (ZS-2) & 40.5 (ZS-1) & 25.8 (ZS-1) & 41.9 (ZS-1) & 32.5 (ZS-2) \\
    Mistral NeMo & 46.2 (ZS-2) & 49.9 (ZS-1) & 43.8 (ZS-2) & 48.6 (ZS-2) & 45.2 (ZS-2) \\
    Phi-3.5 Mini & 34.0 (ZS-2) & 43.7 (ZS-2) & 28.6 (ZS-2) & 43.7 (ZS-2) & 41.1 (ZS-2) \\
    Phi-3.5 MoE & 48.5 (ZS-1) & 46.1 (ZS-1) & 30.5 (ZS-1) & 36.8 (ZS-1) & 30.1 (ZS-2) \\
    \bottomrule
    \end{tabular}
    \label{tab:AMI_table}
\end{table}

\subsection{Used Prompt Templates}
\label{sec:all_prompt_templates}

The tables presented in this section showcase a comprehensive set of prompt templates designed for psychological analysis tasks, meticulously crafted in both English and Arabic to accommodate multilingual applications. These templates are systematically categorized into four primary scenarios: binary classification, multiclass classification, severity assessment, and knowledge evaluation. Each category serves a distinct purpose in assessing various aspects of psychological conditions.

At the core of these templates lies the variable placeholder \textbf{Disorder}, which can be substituted with any specific psychological condition under evaluation, allowing for versatile and targeted assessments. The binary classification templates (ZS-1 [Binary] and ZS-2 [Binary]) are structured to elicit straightforward 'Yes' or 'No' responses, determining the presence or absence of a particular disorder based on the provided social media posts. This simplicity ensures clear and concise data collection, facilitating efficient decision-making processes.

Moving to multiclass classification (ZS-1 [Multiclass] and ZS-2 [Multiclass]), the prompts are designed to identify the most probable disorder from a predefined set of conditions. These templates require respondents to select the disorder that best matches the symptoms presented, or indicate ``No Disorder'' if none are apparent. This approach allows for a more nuanced analysis, capturing a spectrum of psychological conditions beyond binary outcomes.

The severity assessment templates (ZS-1 [Severity] and ZS-2 [Severity]) introduce a quantitative dimension to the evaluation by asking respondents to rate the likelihood or severity of a disorder on a scale from 0 to 1. The label representations within these templates---\textbf{\{0: Disorder representation X\}}, \textbf{\{1: Disorder representation Y\}}, and \textbf{\{N: Disorder representation N\}}---serve as benchmarks for interpreting the severity levels, where ``0'' signifies low severity, ``1'' indicates high severity, and ``N'' denotes the absence of the disorder. This gradation enables a more detailed understanding of the condition's impact.

Lastly, the knowledge evaluation templates (ZS-1 [Knowledge] and ZS-2 [Knowledge]) are tailored to assess the respondent's comprehension and decision-making regarding multiple-choice questions related to psychological disorders. These prompts require the selection of the most appropriate answer without the provision of explanations, ensuring that responses are both accurate and unbiased.

In summary, the prompt templates delineated in the tables offer a structured and adaptable framework for psychological analysis, facilitating precise and scalable evaluations across different classification and assessment scenarios. By leveraging these templates, psychologists and researchers can efficiently gather and interpret data, ultimately contributing to more informed and effective mental health interventions.

\centering
\begin{longtable}[htb]{|c|p{5cm}|p{7cm}|} 
\caption{Prompt templates for psychological analysis tasks in English and Arabic, categorized by binary, multiclass, severity, and knowledge evaluation scenarios.}
\label{tab:11}\\
\toprule 
\textbf{Prompt Template} & 
\textbf{English} & 
\textbf{Arabic} \\ \hline
\midrule
\endfirsthead
\endhead

\textbf{ZS-1 [Binary]}& 
As a psychologist, carefully review the provided social media post(s). Based on the content, do you believe the person is \textbf{\{Disorder\}}? 
(Please respond with 'Yes' or 'No' only, without explaining your reasoning).  
In case of ambiguity: If the post is unclear, choose the most likely label. 
Post(s): 
\textbf{(Inputs)}
&
\begin{RLtext} \small{
بصفتك عالم نفس، قم بمراجعة منشور(ات) وسائل التواصل الاجتماعي المقدمة بعناية. 
استنادًا إلى المحتوى، هل تعتقد أن الشخص\LR{\textbf{\{Disorder\}}}
؟ 
(يرجى الرد بـ نعم أو لا فقط، دون شرح أسبابك). 
في حالة وجود غموض: إذا كان المنشور غير واضح، اختر التسمية الأكثر احتمالاً. 
المنشور(ات): \LR{\textbf{(Inputs)}} }
\end{RLtext}
\\ \hline
\textbf{ZS-2 [Binary]} & [Task]
Analyze the following social media post to determine if the writer exhibits clear symptoms of  \textbf{(Disorder)} according to the provided guidelines.

[Guidelines]
- Concise Response: Respond only with 'Yes' (exhibits clear symptoms of \textbf{(Disorder)}  or 'No' (Does not exhibit clear symptoms of \textbf{(Disorder)}
- No Explanations: Don't provide explanations for your assessment.
- Ambiguity: If the post is unclear, choose the most probable label.
Post(s): \textbf{(Inputs)}& \begin{RLtext} 
[المهمة]
حلل منشور التواصل الاجتماعي التالي لتحديد ما إذا كان الكاتب يُظهر أعراضًا واضحة لـ \LR{\textbf{\{Disorder\}}}
 وفقًا للإرشادات المُقدمة.

[الإرشادات]
- إجابة موجزة: أجب بـ "نعم" فقط (يُظهر أعراضًا واضحة لـ \LR{\textbf{\{Disorder\}}}
 أو "لا" (لا يُظهر أعراضًا واضحة لـ \LR{\textbf{\{Disorder\}}}
.
- عدم تقديم تفسيرات: لا تُقدم تفسيرات لتقييمك.
- الغموض: إذا كان المنشور غير واضح، فاختر التسمية الأكثر احتمالًا.
[المنشور]
\LR{\textbf{(Inputs)}}
\end{RLtext} \\ \hline
\textbf{ZS-1 [Multiclass]} &  &  \begin{RLtext} \small{
بصفتك عالم نفس، قم بمراجعة منشور(ات) وسائل التواصل الاجتماعي المقدمة بعناية.
استنادًا إلى المحتوى، ما هو الاضطراب الأكثر احتمالاً أو الأكثر وضوحًا الذي تعتقد أن الشخص قد يُظهره\LR{\textbf{\{Disorder\}}}
؟
(يرجى ذكر اسم الاضطراب فقط أو "لا يوجد" إذا لم يظهر أي اضطراب).
في حالة وجود غموض: إذا كان المنشور غير واضح، اختر الاضطراب الأكثر احتمالاً.
المنشور(ات):\LR{\textbf{(Inputs)}}}
\end{RLtext} \\ \hline

\textbf{ZS-2 [Multiclass]} & ~ &  \begin{RLtext}
[المهمة]
حلل منشور التواصل الاجتماعي التالي لتحديد ما إذا كان الكاتب يُظهر عرضًا واضحًا لـ \LR{\textbf{\{Disorder\}}}
 وفقًا للإرشادات المُقدمة.

[الإرشادات]
- تحديد الاضطراب: اختر اسم الاضطراب الأكثر احتمالًا والذي يظهر بشكل أكبر من بين \LR{\textbf{\{Disorder\}}}
. إذا لم يكن هناك أي اضطراب واضح، أجب بـ لا يوجد.
- عدم تقديم تفسيرات: لا تُقدم تفسيرات لتقييمك.
- الغموض: إذا كان المنشور غير واضح، فاختر الاضطراب الأكثر احتمالًا.
[المنشور] :\LR{\textbf{(Inputs)}}

\end{RLtext} \\ \hline
\textbf{ZS-1 [Severity]}

& As a psychologist, analyze the provided social media post(s). Is the poster 
\textbf{(Disorder)} 
On a scale of 0 to 1, how likely is this? Label representations:

\textbf{\{0:Disorder representation X\}} 

\vspace{2mm}

\textbf{\{1:Disorder representation Y\}} 

\vspace{2mm}

\textbf{\{N:Disorder representation N\}} 

(Answer with a number between 0 and 1 without explaining your reasoning. If the post is unclear, choose the most probable label without any extra explanation. Please return only the label and no other text.)
    \textbf{(Inputs)} & 
    \begin{RLtext}
بصفتك عالم نفسي، حلل منشور(ات) وسائل التواصل الاجتماعي المقدمة.
هل الشخص الذي كتب المنشور مكتئب؟
على مقياس من \LR{0} إلى \LR{1}, ما مدى ذلك؟
تمثيلات التصنيفات: \vspace{2mm}
\LR{\textbf{\{0:Disorder representation X\}}}

\vspace{2mm}

\LR{\textbf{\{1:Disorder representation Y\}}}

\vspace{2mm}

\LR{\textbf{\{N:Disorder representation N\}}}

(أجب برقم بين \LR{0} و \LR{1} دون شرح الأسباب. إذا كان المنشور غير واضح، اختر التصنيف الأكثر احتمالاً دون أي تفسير إضافي)
(يرجى إرجاع التصنيف فقط دون أي نص آخر)
\LR{\textbf{(Inputs)}} \end{RLtext} \\ \hline
\textbf{ZS-2 [Severity]} &

[Task] Analyze the following social media post(s) to determine the severity of \textbf{(Disorder)} in the writer on a scale from 0 to 1.

[Guidelines]
- Choose a Label: Select a number between 0 and 1 according to the severity of the observed condition.
- Label Representations:
\vspace{2mm}
\textbf{\{0}\textbf{:Disorder representation X\} }

\vspace{2mm}

\textbf{\{1}\textbf{:Disorder representation Y\} }

\vspace{2mm}

\textbf{\{N}\textbf{:Disorder representation N\}  }
- Concise Response: Respond with the number only (no explanation or interpretation).
- Ambiguity: If the post is unclear, choose the most likely label.
[Post] \textbf{(Inputs)}
& \begin{RLtext}
    
[المهمة]
قم بتحليل منشور(ات) وسائل التواصل الاجتماعي التالي لتحديد مدى شدة مكتئب لدى الكاتب على مقياس من\LR{0} الي \LR{1} 
[الإرشادات]
- اختر تصنيفًا: حدد رقمًا بين\LR{0} و \LR{1}وفقًا لمدى شدة الحالة الظاهرة.
- تمثيلات التصنيفات: 
 \vspace{2mm}
\textbf{\LR{\textbf{\textbf{\{0:disorder representation X\}}}}}

\vspace{2mm}

\textbf{\LR{\textbf{\textbf{\{1:disorder representation Y\}}}}}

\vspace{2mm}

\textbf{\LR{\textbf{\{N:disorder representation N\}}} }
إجابة موجزة: أجب بالرقم فقط (دون شرح أو تفسير).
- الغموض: إذا كان المنشور غير واضح، اختر التصنيف الأكثر احتمالاً.

[المنشور]
\LR{\textbf{(Inputs)}} \end{RLtext} \\ \hline
\textbf{ZS-1 [Knowledge]} & As a psychologist, please analyze the following multiple-choice question and select the most appropriate answer.
Do not provide any explanation for your choice.
Indicate your choice using a single English letter.
(Provide only one answer)
\textbf{(Inputs)} 
&
\begin{RLtext}
بصفتك عالم نفس، يرجى تحليل سؤال الاختيار من متعدد التالي واختيار الإجابة الأكثر ملاءمة.
لا تقدم أي تفسير لاختيارك.
أشر إلى اختيارك بحرف عربي واحد.
(قدم إجابة واحدة فقط)
\LR{\textbf{(Inputs)}} \end{RLtext}\\ \hline

\textbf{ZS-2 [Knowledge]} & [Task]
Analyze the following multiple-choice question and determine the most appropriate answer.

[Guidelines]
- Select only one answer.
- Do not provide any explanations for your choice.
- Indicate your choice using the corresponding English letter for the answer.

[Question]
\textbf{(Inputs)} & \begin{RLtext} [المهمة]
قم بتحليل سؤال الاختيار من متعدد التالي وتحديد الإجابة الأكثر ملاءمة.

[الإرشادات]
- اختر إجابة واحدة فقط.
- لا تُقدم أي تفسيرات لاختيارك.
- أشر إلى اختيارك باستخدام الحرف العربي الموافق للإجابة.

[السؤال]
\LR{(\textbf{Inputs})} \end{RLtext}\\ \hline
\end{longtable}

\end{document}